\documentclass[journal,10pt,twocolumn,compsoc]{IEEEtran}

\newif\ifdouble
\doubletrue

\usepackage[utf8x]{inputenc}
\usepackage[T1]{fontenc}
\usepackage[]{standalone}
\usepackage{tikz,pgfplots}
\usepackage{graphicx}
\usepackage{grffile}
\usepackage{xcolor,colortbl}
\usepackage[caption=false]{subfig}
\usepackage{amsmath}
\usepackage{booktabs}
\usepackage{multirow}
\usepackage{diagbox}
\usepackage{amsthm}
\usepackage{amssymb}
\usepackage{amsfonts}
\usepackage{wasysym}
\usepackage{algorithm}
\usepackage{algorithmic}
\usepackage{balance}

\usepackage[breaklinks=true,letterpaper=true,colorlinks,bookmarks=false]{hyperref}

\usepackage{url}

\catcode`~=11

\catcode`~=13

\usepackage[numbers,sort&compress]{natbib}

\usetikzlibrary{arrows,automata,calc,shapes,shapes.symbols,positioning,fit,shadows}
\usetikzlibrary{pgfplots.statistics, pgfplots.colorbrewer} 
\usepgfplotslibrary{groupplots}

\pgfplotsset{
    compat=1.12,
    every legend to name picture/.style={scale=0.7},
    every legend to name picture/.style={font=\scriptsize},
}

\usepackage{etoolbox}
\makeatletter
\patchcmd\@combinedblfloats{\box\@outputbox}{\unvbox\@outputbox}{}{\errmessage{\noexpand patch failed}}
\makeatother

\definecolor{mycolor1}{rgb}{0.85,0.85,1.0}
\definecolor{mycolor2}{rgb}{1.0,0.85,0.85}
\definecolor{mycolor3}{rgb}{0.0,0.75,0.75}%
\definecolor{mycolorg}{rgb}{0.5,1.00,0.5}
\definecolor{mycolorr}{rgb}{1.0,0.5,0.5}
\definecolor{mycolorm}{rgb}{1.0,0.0,1.0}%
\definecolor{green}{rgb}{0.0,0.5,0.0}
\definecolor{lightgreen}{rgb}{0.0,1.0,0.0}

\definecolor{color0}{rgb}{0.96,0.96,0.98}

\definecolor{color3}{rgb}{0.333333333333333,0.658823529411765,0.407843137254902}
\definecolor{color1}{rgb}{0.298039215686275,0.447058823529412,0.690196078431373}
\definecolor{color2}{rgb}{0.866666666666667,0.517647058823529,0.32156862745098}
\definecolor{color4}{rgb}{0.768627450980392,0.305882352941176,0.32156862745098}

\tikzset{
    >=stealth',
    line/.style = {draw, ->},
    narline/.style = {text width=2cm, font = \small},
    operation/.style={
           rectangle,
           rounded corners,
           draw=black,
           drop shadow={shadow scale=0.95},
           fill=green!10,
           minimum width=1.4cm,
           minimum height=1.1cm,
           text width=1.65cm,
           text centered},
    flow/.style={
           diamond,
           aspect=2,
           draw=black, thick,
           fill=blue!10,
           minimum height=1em,
           text centered},
    rect/.style={
           rectangle,
           draw=black, thick,
           minimum height=1em,
           text centered},
    input/.style={
           rectangle,
           draw=black, thick,
           minimum height=1em,
           text centered}
}

\title{Neural Imaging Pipelines - the Scourge or Hope of Forensics?}

\author{
  Paweł Korus and Nasir Memon,~\IEEEmembership{Fellow,~IEEE}
  \IEEEcompsocitemizethanks{\IEEEcompsocthanksitem P. Korus is with the Tandon School of Engineering, New York University, USA, and also with the Department of Telecommunications, AGH University of Science and Technology, Poland (e-mail: pkorus@nyu.edu).}
\IEEEcompsocitemizethanks{\IEEEcompsocthanksitem N. Memon is with the Tandon School of Engineering, New York University, USA(e-mail: memon@nyu.edu).}
  \vspace*{-0.5cm}
}

\begin{document}

\IEEEtitleabstractindextext{
\begin{abstract}
Forensic analysis of digital photographs relies on intrinsic statistical traces introduced at the time of their acquisition or subsequent editing. Such traces are often removed by post-processing (e.g., down-sampling and re-compression applied upon distribution in the Web) which inhibits reliable provenance analysis. Increasing adoption of computational methods within digital cameras further complicates the process and renders explicit mathematical modeling infeasible. While this trend challenges forensic analysis even in near-acquisition conditions, it also creates new opportunities. This paper explores end-to-end optimization of the entire image acquisition and distribution workflow to facilitate reliable forensic analysis at the end of the distribution channel, where state-of-the-art forensic techniques fail. We demonstrate that a neural network can be trained to replace the entire photo development pipeline, and jointly optimized for high-fidelity photo rendering and reliable provenance analysis. Such optimized \emph{neural imaging pipeline} allowed us to increase image manipulation detection accuracy from approx. 45\% to over 90\%. The network learns to introduce carefully crafted artifacts, akin to digital watermarks, which facilitate subsequent manipulation detection. Analysis of performance trade-offs indicates that most of the gains can be obtained with only minor distortion. The findings encourage further research towards building more reliable imaging pipelines with explicit provenance-guaranteeing properties. 
\end{abstract}

}

\maketitle

\IEEEdisplaynontitleabstractindextext
\IEEEpeerreviewmaketitle

\section{Introduction}
\label{sec:introduction}

\begin{figure*}[!t]
    \includegraphics[width=1.0\textwidth]{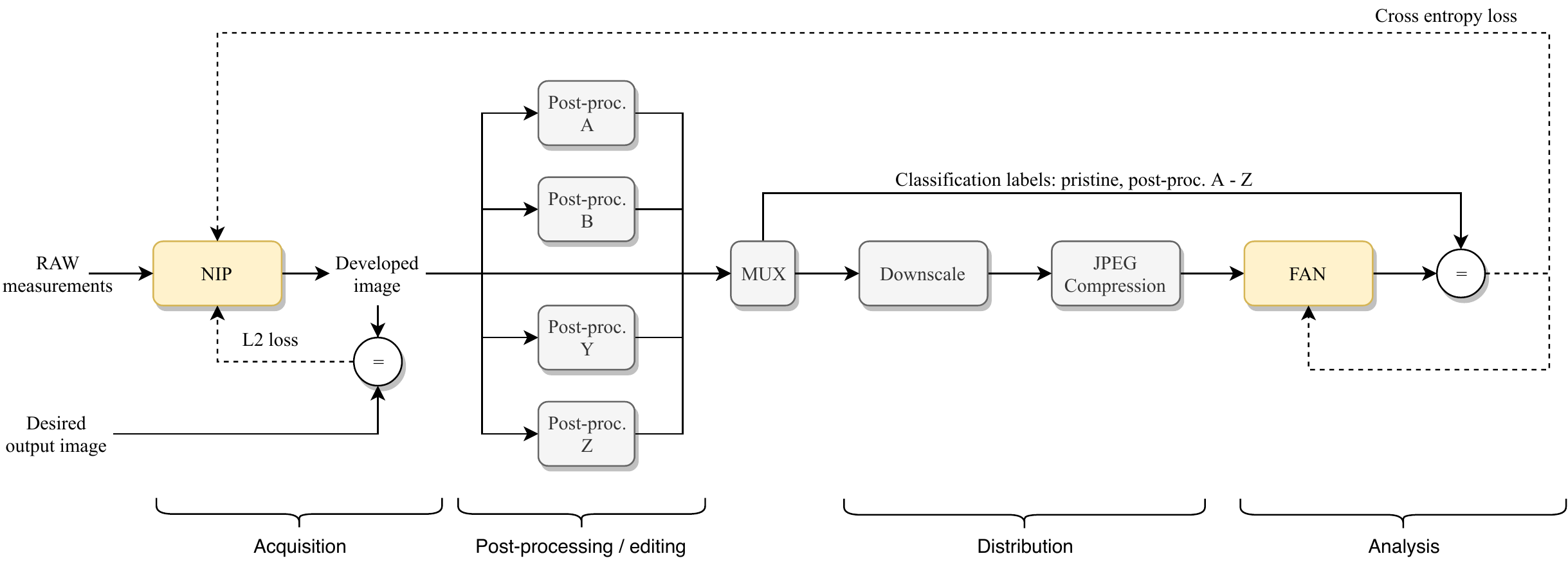}
    \caption{Optimization of the image acquisition and distribution channel to facilitate photo provenance analysis. The neural imaging pipeline (NIP) is trained to develop images that both resemble the desired target images, but also retain meaningful forensic clues at the end of complex distribution channels.}
    \label{fig:nip-training-protocol}
\end{figure*}

Assessment and protection of the integrity of digital photographs is one of the most challenging and important problems in multimedia communications. Photographs are commonly used for documentation of important events, and as such, require efficient and reliable authentication protocols. Our media acquisition and distribution workflows are built with entertainment in mind, and not only fail to provide explicit security features, but actually work against them. Image compression standards exploit heavy redundancy of visual signals to reduce communication payload, but optimize for human perception alone. Security extensions of popular standards~\cite{jpegSecurity} do not address modern distribution requirements, like re-compression or device adaptation, and lack in adoption. 

Two main approaches to photo content authentication include~\cite{korus2017digital,piva2013overview,stamm2013information}: (1) pro-active protection based on digital signatures or watermarking; (2) passive forensic analysis that exploits statistical traces introduced by the acquisition pipeline or subsequent editing. While pro-active solutions provide superior performance and protection features (e.g., precise tampering localization~\cite{yan2017multi}, or reconstruction of tampered content~\cite{Korus2014b,korus2014iterative}), it requires explicit generation of protected image copies. Such functionality should be integrated in digital cameras~\cite{blythe2004secure}, but vendors lack incentives to introduce and support such mechanisms. Registration of photographs and their signatures in a central repository~\cite{lin2012image} works for limited deployments, such as law-enforcement/public safety applications~\cite{motorola}. Several companies and researchers are currently experimenting with blockchain-based repositories~\cite{keeex,ewitness,truepic}.

Passive forensics, on the other hand, relies on our knowledge of the photo acquisition pipeline, and statistical artifacts introduced by its successive steps. While this approach is well suited for analyzing any digital photograph, it often falls short due to complex post-processing. Digital images are not only heavily compressed, but also enhanced or even manipulated before, during or after dissemination. Popular images have many online incarnations, and tracing their distribution and evolution has spawned a new field of image phylogeny~\cite{dias2013large,dias2013toward} which relies on visual differences between multiple images to infer their relationships and editing history. However, phylogeny does not provide any tools to reason about the authenticity or history of individual images. As a result, reliable authentication of real-world online images remains untractable~\cite{zampoglou2016large}.

At the moment, forensic analysis often yields useful results in near-acquisition scenarios. Analysis of native images straight from the camera is more reliable, and even seemingly benign implementation details, like rounding operators in image signal processors~\cite{agarwal2017photo}, provide useful clues. Unfortunately, most forensic traces quickly become unreliable with post-processing. One of the most reliable forensic tools involves analysis of the imaging sensor's artifacts (the photo response non-uniformity, PRNU) which can be used both for source attribution and content authentication problems~\cite{Chen2008}.

Rapid progress in computational imaging will soon challenge forensics even in near-acquisition authentication. In the pursuit of better image quality and convenience, digital cameras (especially smartphones) employ sophisticated post-processing, and final photographs cease to resemble the original signals captured by the sensor(s). Adoption of machine learning has recently challenged many long-standing limitations of digital photography, such as: high-quality low-light photography~\cite{Chen2018,night-sight-pixel3}; single-shot HDR with overexposed content recovery~\cite{Eilertsen2017}; high-quality digital zoom from multiple shots~\cite{super-resolution-pixel3}; enhancement of smartphone-captured images with weak supervision from DSLR photos~\cite{ignatov2017wespe}.

These remarkable results demonstrate tangible benefits of replacing the entire acquisition pipeline with neural networks. Hence, it will be necessary to assess the impact of \emph{neural imaging pipelines} on existing forensic protocols. While important, such evaluation can be seen as damage assessment and control rather than a solution for the future. We believe it is imperative to consider new opportunities for security-oriented design of cameras and multimedia dissemination channels that come with adoption of neural imaging processors. 

In this paper, we propose to optimize neural imaging pipelines to improve photo provenance analysis capabilities in complex distribution channels. We exploit end-to-end optimization of the entire photo acquisition and distribution channel to ensure that reliable authentication decisions can be made even after complex post-processing, where classical forensics fails (Fig.~\ref{fig:nip-training-protocol}). We believe the ongoing revolution in camera design creates a unique opportunity to address the limitations of our current technology. While adoption of authentication watermarking was stifled by the necessity to modify camera design, our approach exploits the flexibility of neural networks to learn relevant integrity-preserving features within the expressive power of the model. With solid understanding of neural imaging pipelines, and a rare opportunity of replacing the well-established and security-oblivious technology, we have a chance to significantly improve photo authentication capabilities in next-generation devices. 

We aim to inspire discussion about novel, learnable camera designs that could improve photo provenance analysis capabilities. The capability of rapidly learning the entire imaging pipeline for novel hardware components~\cite{Jiang2017,Syu2018} opens new perspectives for security-oriented enhancements. We demonstrate that it is possible to optimize the pipeline to significantly improve detection of photo manipulation at the end of a complex real-world distribution channel, where state-of-the-art deep-learning techniques fail. The main contributions of our work include:
\begin{enumerate}
    \itemsep0em
    \item The first end-to-end optimization of the imaging pipeline with explicit photo provenance objectives;
    \item The first security-oriented discussion of neural imaging pipelines and the inherent trade-offs;
    \item Significant improvement of forensic analysis performance in challenging, heavily post-processed conditions;
    \item First discussion of the impact of computational imaging on the stability of conventional forensics protocols.
\end{enumerate}

This work is an extension of our conference paper~\cite{korus2018content}. The paper is organized as follows. First, we review related work on computational imaging, application of machine learning to imaging pipeline design, as well as conventional methods of photo authentication (Section~\ref{sec:related-work}). Then, we discuss the application of neural imaging pipelines, and assess their impact on classical forensics analysis (Section~\ref{sec:pipelines}). The following sections explain the proposed pipeline modeling and optimization approach (Section~\ref{sec:pipeline-optimization}) and present the results of experimental evaluation (Section~\ref{sec:secure-nip-results}). Finally, we conclude and discuss the perspectives for future research (Section~\ref{sec:conclusions}).

To facilitate further research in this direction, and enable reproduction of our results, our neural imaging toolbox is available online at \url{https://github.com/pkorus/neural-imaging}.

\section{Related Work}
\label{sec:related-work}

In this section, we review recent trends in computational photography, including imaging pipeline design, and approximation of image processing operators with artificial neural networks. We also discuss existing techniques and research trends in image authentication, including both watermarking and forensics-based solutions.

\subsection{Imaging Pipelines and Processors} 

Learning individual steps of the imaging pipeline has a long history~\cite{kapah2000demosaicking} but regained momentum in recent years thanks to deep learning. Naturally, the research focused on the most difficult operations, i.e., demosaicing~\cite{Gharbi2016,Tan2017,Kokkinos2018,Syu2018} and denoising~\cite{burger2012image,zhang2017beyond,lehtinen2018noise2noise}. New techniques delivered not only better performance, but also additional features. Gharbi et al. proposed a convolutional neural network (CNN) architecture for joint demosaicing and denoising~\cite{Gharbi2016}. Syu et al. used CNNs for joint optimization of the color filter array and a corresponding demosaicing filter~\cite{Syu2018}. Researchers are also training networks to \emph{unprocess} RGB images to RAW inputs, which aims to obtain more realistic training data, imperative in high-quality denoising~\cite{brooks2018unprocessing}.

Optimization efforts have recently expanded to a complete replacement of the pipeline with learned components. The L3 model by Jiang et al. approximates the entire photo development process with a large collection of local linear filters~\cite{Jiang2017}. The model aims to facilitate research and development efforts for non-standard camera designs, like custom RGBW (red-green-blue-white) and RGB-NIR (red-green-blue-near-infra-red) color filter arrays. The FlexISP model can be optimized for various camera designs and image formation models~\cite{heide2014flexisp}. Optimization objectives can include not only standard metrics like demosaicing or denoising performance, but also inverse problems like camera shake and out-of-focus blur reduction. Chen et al. trained a UNet model~\cite{ronneberger2015u} to develop high-quality photographs in low-light conditions~\cite{Chen2018} by exposing it to paired examples of images taken with short and long exposure. The network learned to develop high-quality well-exposed color photographs from underexposed raw input, and yielded better performance than traditional image post-processing based on brightness adjustment and denoising. Eilertsen et al. also used a UNet model to develop high-dynamic range images from a single shot~\cite{Eilertsen2017}. The network  learned not only tone mapping, but also recovery of overexposed highlights. This significantly simplifies HDR photography by eliminating bracketing and ghosting artifacts. 

In contrast to mainstream research which prioritizes image quality, adoption of neural networks allows to optimize not only for human perception, but also for machine consumption. Buckler et al. prioritize low-power processing given machine perception objectives~\cite{buckler2017reconfiguring}. A company Algolux~\cite{algolux} optimizes imaging pipelines for object detection in autonomous vehicles.

Machine learning can also simplify image fusion, which is increasingly adopted in photography~\cite{wang2019stereoscopic,hasinoff2016burst,night-sight-pixel3,light-l16}. Hasinoff et al. proposed a burst pipeline, which takes multiple short-exposure photographs and combines them to improve low-light and HDR performance~\cite{hasinoff2016burst}. Their pipeline uses a simple pairwise filter, which successively merges new frames into a reference image. To eliminate ghosting on moving objects, fast sub-pixel alignment is used. Building upon this technique, the most recent smartphones seek improvements in extremely low-light conditions~\cite{night-sight-pixel3}. A recent digital camera (Light L16~\cite{light-l16}) uses 16 lenses and sensors with different configurations to provide high-quality imaging capabilities within a small form factor. The camera activates various groups of sensors depending on capture settings, and relies on a proprietary image fusion engine to develop the photographs and deliver advanced features, e.g., post-capture depth of field control.

\subsection{Image Enhancement and Processing}

CNN architectures are also considered for image enhancement and processing. Ignatov et al. developed a system for smartphone photo quality enhancement~\cite{ignatov2017dslr}. The authors captured synchronized pairs of images from low-cost mobile cameras, and high-quality DSLRs. Then, they trained a neural network to regress high-quality photos given low-quality input. In a follow up work, they extended the approach to dispense with strong supervision. They adopted an image translation-based approach which yields competitive results despite training on unpaired low-quality and high-quality photo collections~\cite{ignatov2017wespe}. 

Automatic photo adjustment using neural networks is another active research topic~\cite{yan2016automatic,liu2019learning}. Deep networks were shown to be effective not only in subjective style enhancement, but also in fast approximation of complex image processing operators, e.g., tone mapping, or dehazing~\cite{Chen2017}. Recent work by Gharbi et al. suggests that thanks to complex feature detection capabilities, deep architectures can learn not only global processing operators, but also subtle, contextual adjustments, e.g., brightening of human faces~\cite{gharbi2017deep}.

A recent trend in post-processing involves simulation of shallow depth-of-field in low-cost smart-phone photography. The most successful systems combine face location, semantic image segmentation, and depth information to synthesize a convincing blur of the scene background~\cite{wadhwa2018synthetic}. Depth information can be estimated using either multi-camera setups, parallax from subtle camera movements, or dual-pixel auto-focus hardware. Synthesis methods range from simple composition models up to approximations of light field processing.

\subsection{Digital Image Forensics} 

\begin{figure*}[!t]
    \centering
    \includegraphics[width=0.95\textwidth]{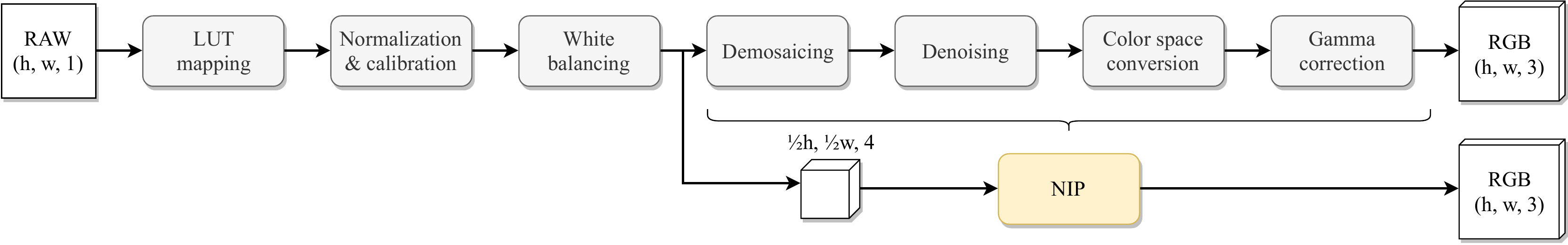}
    \vspace{0.0cm}
    \caption{Adoption of a neural imaging pipeline to develop color RGB images from raw sensor measurements: (top) the standard imaging pipeline; (bottom) the neural imaging pipeline.}
    \label{fig:pipeline}
\end{figure*}

Forensic analysis addresses questions regarding photo provenance and processing history, and considers either \emph{physical integrity} (consistency of lighting, shadows, perspective, etc) or \emph{digital integrity} (consistency of low-level signal features)~\cite{piva2013overview,korus2017digital}. The former are typically manual, or semi-automatic, as they require good high-level understanding of image content and laws of physics. The latter techniques are automatic, and rely on intrinsic statistical traces introduced at the time of photo acquisition (or manipulation). Based on our knowledge of the standard imaging pipeline, researchers have proposed many techniques for problems ranging from source attribution to content authentication. The initial research emphasized formal mathematical modeling of image processing operators and corresponding detectors. Increasingly complex post-processing forced a transition to data-driven techniques. However, recent evaluation revealed limited applicability of both approaches on real-world data~\cite{zampoglou2016large}.

The current research focuses on three main directions: (1) learning deep features relevant to low-level forensic analysis for problems like manipulation detection~\cite{bayar2018constrained,zhou2018,bondi2017tampering}, identification of the social network of origin\cite{amerini2017tracing}, camera model identification~\cite{cozzolino2018camera,cozzolino2018noiseprint,bondi2017first}, or detection of artificially generated content~\cite{li2018detection,marra2018gans,guera2018deepfake,rossler2019faceforensics}; (2) adoption of high-level vision to automate manual analysis that exposes physical inconsistencies, such as reflections~\cite{sun2017object,wengrowski2017reflection}, or shadows~\cite{kee2014exposing}; (3) addressing counter-forensic and adversarial attacks~\cite{barni2018adversarial}.

The bulk of forensics research assumes a standard imaging pipeline, which increasingly ceases to hold due to increasing adoption of computational methods. Early symptoms are visible even in older devices, where image stabilization hampers attribution and authentication of video recordings~\cite{taspinar2016source,mandelli2018facing}. We expect that such problems will quickly deteriorate, and become pervasive even for static photography which starts to rely on advanced image fusion techniques (such as the burst imaging pipeline~\cite{hasinoff2016burst}, and multi-sensor setups~\cite{light-l16}).

To the best of our knowledge, there are currently no efforts to assess the impact of new pipelines on existing forensics protocols. We are also not aware of any studies addressing the sensitivity of forensics models to the use of different pipelines (e.g., different RAW development software). Recent research in image steganography argues that even the same pipeline with different acquisition settings (e.g., ISO sensitivity) can be considered a different source with distinct statistical properties \cite{bas2016steganography,gibouloto2018steganalysis}. While this observations awaits a rigorous study in the forensics setting, our results with PRNU analysis seem to support it (Section~\ref{sec:impact-on-forensics}).

\subsection{Digital Watermarking}

Adoption of watermarking for content authentication requires two modules: an \emph{encoder} which generates protected versions of the photographs (embeds a watermark); and a \emph{decoder} which analyzes the embedded watermark and makes a decision about content authenticity~\cite{korus2017digital}. The watermark is typically designed to be \emph{semi-fragile}, i.e., it remains in the image when legitimate post-processing is applied (e.g., compression, or brightness adjustments), but is destroyed by malicious editing (e.g., content forgery). As a result, the watermark is typically embedded by modulating middle frequencies, which have limited impact on image quality but remain robust to post-processing. 

Authentication watermarks allow for binary decisions. The detector measures the presence of the expected watermark and deems the image (or its fragment) authentic if the watermark is detected, or tampered if no watermark is found. It is currently not possible to infer image editing history based on the extracted watermark. Such functionality was envisioned by \emph{telltale} watermarking, but hand-crafting watermarks to facilitate such decisions turned out to be infeasible. We are aware of only toy examples, e.g., a recent study designed a visible watermarking scheme for counting of JPEG compression steps~\cite{carnein2016telltale}. The scheme works by placing in the image small blocks with carefully-crafted convergence properties (stabilization of pixel changes with successive re-compressions).

Current research explores neural network architectures for information hiding, but at the current stage addresses simple models for steganography~\cite{baluja2017hiding,tang2019cnn}. A recent framework allows to balance embedding objectives and optimize either for secrecy (steganography) or robustness (watermarking)~\cite{zhu2018hidden}. We are not aware of any efforts exploring neural network-based optimization of authentication watermarks.

\section{Neural Imaging Pipelines}
\label{sec:pipelines}

\begin{figure*}
    \centering
    \includegraphics[width=\textwidth]{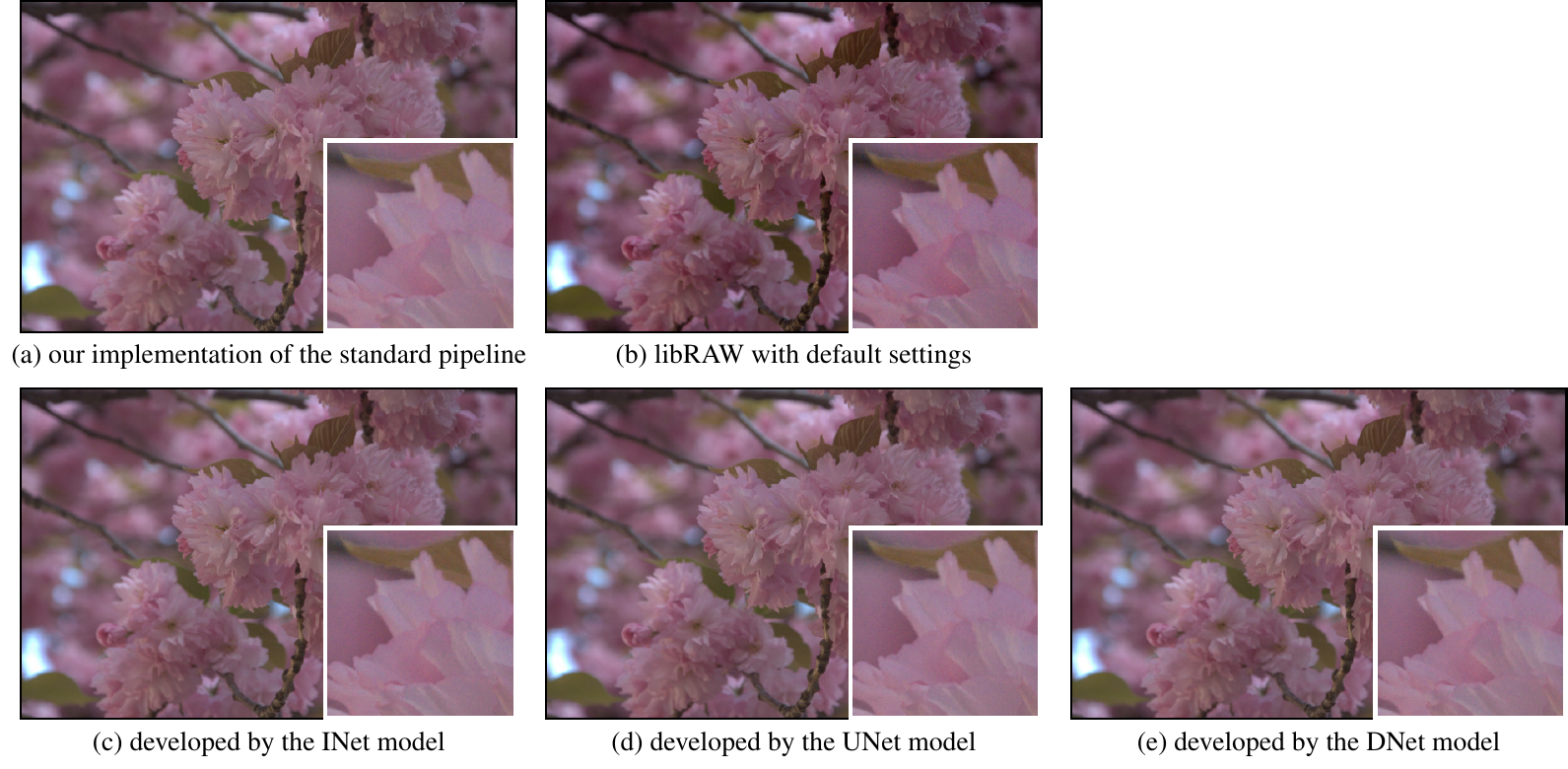}
    \caption{An example full-resolution (12.3~Mpx) image developed with standard pipelines (ab) and the considered NIPs (cde): image \emph{r23beab04t} shot with Nikon D90 (Raise dataset~\cite{dataset:raise}). (Full-resolution images are included as JPEGs (quality 85, 4:2:0) to limit PDF size; close-ups are included as lossless PNGs.)}
    \label{fig:nip-full-resolution-dev}
\end{figure*}

This section reviews the standard imaging pipeline, and details how it can be replaced with its neural counterpart. We then present the results of training 3 different neural pipelines for several camera models. Finally, we discuss the impact of the replacement on established forensics protocols.

\subsection{The standard imaging pipeline}

The imaging pipeline is responsible for converting raw sensor measurements into color images ready for display. A typical pipeline (Fig.~\ref{fig:pipeline}) will contain the following steps: 1) \emph{normalization and calibration} which involves subtracting the black level and division by the saturation value; 2) \emph{white balancing} which involves global adjustment of  the intensities of color components that ensures correct color perception; 3) \emph{demosaicing} which interpolates color components at unmeasured locations; 4) \emph{denoising} which removes grain, especially at higher ISO settings; 5) \emph{color space conversion} from camera RGB to sRGB which involves a linear matrix multiplication; 6) \emph{gamma correction} which expands darker tones to account for logarithmic perception of the human vision system. More advanced pipelines often contain additional steps for sophisticated tone-mapping, masking of sensor defects, or immediate quality enhancement. 

In our study, we implemented a standard pipeline in Python and used it as the optimization target for our neural networks. We used \emph{rawkit}~\cite{rawkit} wrappers over \emph{libRAW}~\cite{libraw} to access sensor measurements and meta-data in RAW photographs. \emph{Rawkit} automatically inverts the non-linear mapping of pixel intensities employed by some vendors (the LUT parameters are available in image meta-data). Demosaicing was performed using an adaptive algorithm by Menon et al.~\cite{menon2007demosaicing}. Since most images in our data sets were shot in good lighting conditions, we did not include denoising. Due to patch-based neural network training, our pipeline does not use any post-processing that requires global knowledge about the image, e.g., automatic exposure correction. An example full-resolution image developed by our Python pipeline is shown in Fig.~\ref{fig:nip-full-resolution-dev}a.

\subsection{The neural imaging pipeline}
\label{sec:nip-training}

We replace the entire imaging pipeline with a CNN model which develops raw sensor measurements into color RGB images (Fig.~\ref{fig:pipeline}). The input images are pre-process by reversing the nonlinear value mapping according to the camera's LUT, subtracting black levels from the edge of the sensor, normalizing by sensor saturation values, and applying white-balancing according to shot settings. We also re-organized the inputs by reshaping them into RGGB feature maps of size ($\frac{h}{2}, \frac{w}{2}, 4$) where $(w,h)$ are the width/height of the full-resolution raw Bayer image. This ensures well-standardized inputs with values in [0, 1]. 

\begin{table}[t]
\caption{Considered neural imaging pipelines}
\label{tab:pipelines}
\centering
\begin{footnotesize}    
    \begin{tabular}{lccc}
        \toprule 
        & \textbf{INet} & \textbf{UNet} & \textbf{DNet} \tabularnewline
        \midrule
        \textbf{\# Parameters} & 321 & 7,760,268 & 493,976\tabularnewline
        \textbf{PSNR [dB]} & 42.8 & 44.3 & 46.2\tabularnewline
        \textbf{SSIM} & 0.989 & 0.990 & 0.995\tabularnewline
        \textbf{Train. speed [it/s]} & 8.80 & 1.75 & 0.66\tabularnewline
        \textbf{Train. time} & 17 - 26 min & 2-4 h & 12 - 22 h\tabularnewline
        \bottomrule
    \end{tabular}
\end{footnotesize}
\end{table}

\begin{figure}
    \centering
    \includegraphics[width=1.0\columnwidth]{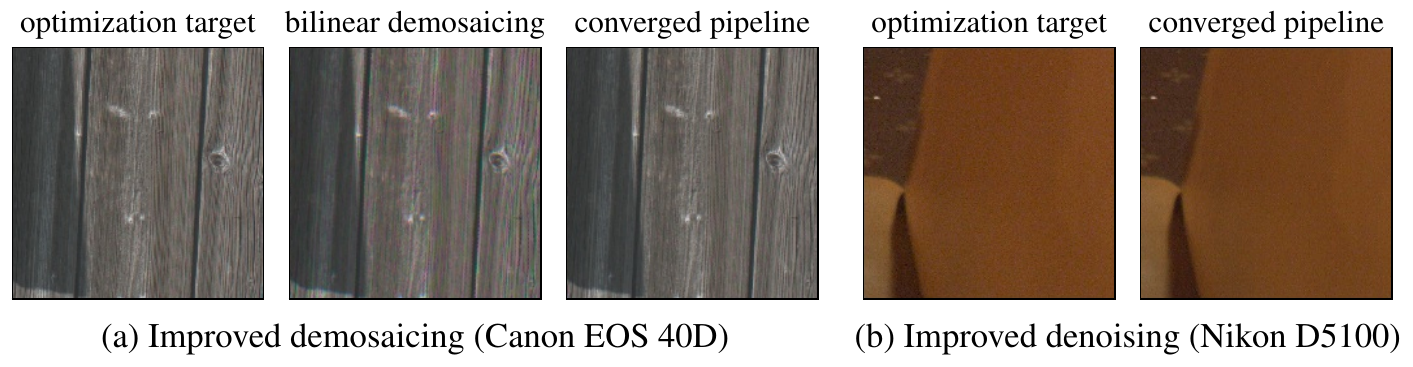}
    \caption{Examples of serendipitous image quality improvements obtained by neural imaging pipelines: (a) better demosaicing; (b) better denoising.}
    \label{fig:nip-visual-improvement}
\end{figure}

\begin{figure*}[!t]
    \centering
    \includegraphics[width=1.00\textwidth]{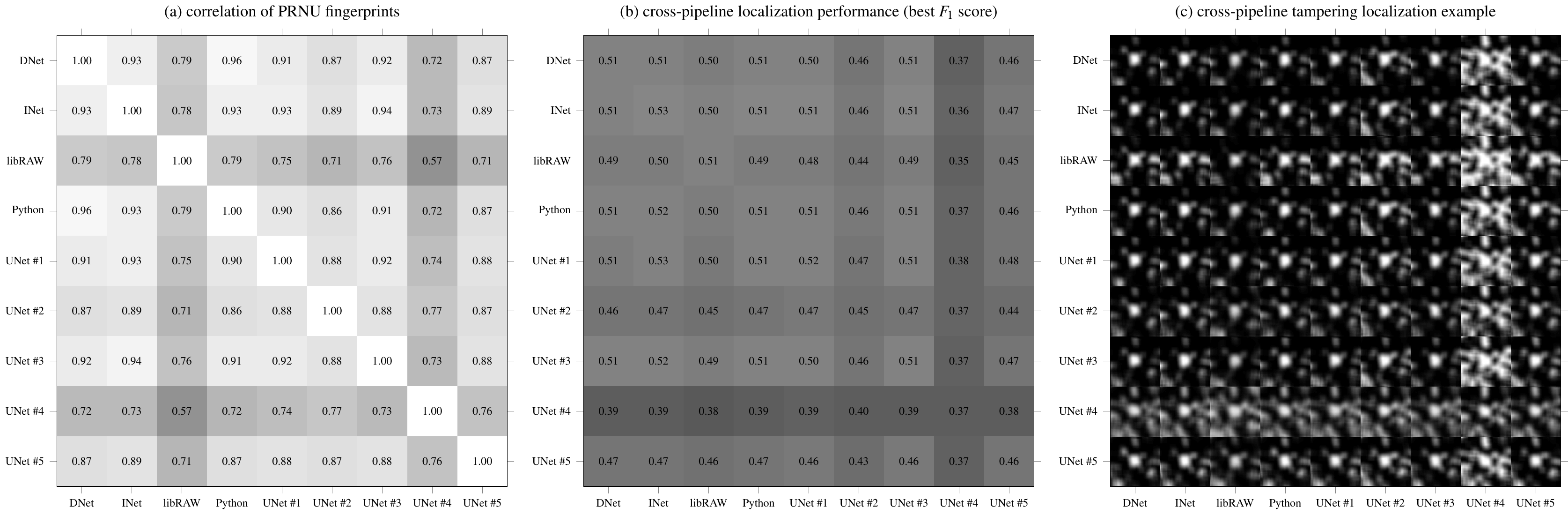}
    \caption{Impact of imaging pipeline on PRNU-based forensics analysis: (a) correlation between PRNU fingerprints obtained from the same images developed by different imaging pipelines; (b) cross-pipeline tampering localization performance (best $F_1$ scores); (c) cross-pipeline tampering localization example with an attempt to localize a horizontally flipped square region in the middle (forgery size $256\times256$~px; analysis window size $129\times129$~px).}
    \label{fig:prnu}
\end{figure*}

We considered 3 NIP models with various complexity and design principles (Tab.~\ref{tab:pipelines}; see supplement for detailed architecture). \emph{INet} is a simple convolutional network with layers corresponding to successive steps of the standard pipeline. \emph{UNet} represents the popular UNet architecture~\cite{ronneberger2015u} adapted from~\cite{Chen2018}. \emph{DNet} is an adaptation of a recent model proposed for joint demosaicing and denoising~\cite{Gharbi2016}. All considered networks are fully convolutional, and work on inputs of arbitrary size. We take advantage of this property during training, when we feed randomly sampled small patches instead of full-size images. This significantly simplifies training, and the networks can be used on full-resolution inputs without any changes.

We collected a data-set with RAW images from 8 cameras (Tab.~\ref{tab:cameras}). The photographs come from two public (Raise~\cite{dataset:raise} and MIT-5k~\cite{dataset:fivek}) and one private data-sets. For each camera, we randomly selected 150 images with landscape orientation. For training, we used 120 full-resolution images, which were sampled in each iteration to yield batches of 20 examples with spatial resolution $128\times128$~px. For validation, we used a fixed set of $512\times512$~px patches extracted from the remaining 30 images. The models were trained to reproduce images developed by our Python pipeline (separate model for each camera), and were penalized by the $L_2$ loss averaged over all dimensions of the output tensor (with patch size $p$):
\begin{equation}
    L_{\text{nip}} = \frac{1}{3p^2N} \sum_{n} \| y_n - \text{nip}(x_n~|~\theta_{\text{nip}}) \|_2 
\end{equation}
\noindent where: $\theta_{\text{nip}}$ are the parameters of the NIP network; $x_n$ are the raw sensor measurements for the $n$-th example patch ($1\times\frac{p}{2}\times\frac{p}{2}\times4$ tensor); $\text{nip}(x_n)$ is the color RGB image developed by the NIP from $x_n$ ($1\times p \times p \times 3$ tensor).

\begin{table}[t]
\caption{Digital cameras used in our experiments}
\label{tab:cameras}
\centering
\begin{footnotesize}
\begin{tabular}{lrrrr}
\toprule 
\textbf{Camera} & \textbf{Sensor} & \#\textbf{Images}$^1$ & \textbf{Source} & \textbf{Bayer}\tabularnewline
\midrule
Canon EOS 5D  & 12 Mpx & 864 dng & MIT-5k & RGGB\tabularnewline
Canon EOS 40D & 10 Mpx & 313 dng & MIT-5k & RGGB\tabularnewline
Nikon D5100   & 16 Mpx & 288 nef & Private & RGGB\tabularnewline
Nikon D700    & 12 Mpx & 590 dng & MIT-5k & RGGB\tabularnewline
Nikon D7000   & 16 Mpx & $>$1k nef & Raise & RGGB\tabularnewline
Nikon D750    & 24 Mpx & 312 nef & Private & RGGB\tabularnewline
Nikon D810    & 36 Mpx & 205 nef & Private & RGGB\tabularnewline
Nikon D90     & 12 Mpx & $>$1k nef & Raise & GBRG\tabularnewline
\bottomrule
\multicolumn{5}{l}{\footnotesize $^1$ RAW file formats: nef (Nikon); dng (generic, Adobe)}
\end{tabular}
\end{footnotesize}
\end{table}

All NIPs successfully reproduced target images with high fidelity. Fig.~\ref{fig:nip-full-resolution-dev} shows an example full-resolution (12.3~Mpx) image developed with: (a) our standard Python pipeline; (b) libRAW with default settings; (cde) the considered \emph{INet}, \emph{UNet} and \emph{DNet} models. The resulting color photographs are visually indistinguishable from the targets. Objective fidelity measurements for the validation set are collected in Tab.~\ref{tab:pipelines} (average over all cameras). Interestingly, the trained models often revealed better denoising and demosaicing performance (Fig.~\ref{fig:nip-visual-improvement}), despite the lack of denoising in the simulated pipeline, and the lack of explicit optimization objectives. 

Of all of the considered models, \emph{INet} was the easiest to train - not only due to its simplicity, but also because it could be initialized with meaningful parameters that already produced reasonably close results. We initialized the demosaicing filters with bilinear interpolation, color space conversion with an example multiplication matrix, and gamma correction with a toy model separately trained to reproduce this non-linearity. \emph{UNet} was initialized randomly, but improved rapidly thanks to skip connections. \emph{DNet} took the longest and for a long time had problems with faithful color rendering. The typical training times are reported in Tab.~\ref{tab:pipelines} (measured on Nvidia Tesla K80 GPUs). The models were trained until the relative change of the average validation loss for the last 5 epochs dropped below $10^{-4}$. The maximum number of epochs was 50,000. For \emph{DNet} we adjusted the stopping criterion to $10^{-5}$ due to slow training that sometimes terminated prematurely with incorrect color rendering.

\subsection{Impact on Classical Forensics}
\label{sec:impact-on-forensics}

In this experiment, we take advantage of multiple available pipelines and assess their impact on classical image forensics. We use PRNU fingerprint analysis~\cite{Chen2008} (one of the most reliable forensic techniques) which characterizes the imaging sensor - a component of the camera which remains unchanged across our experiments. Our evaluation is divided into two parts with neural pipelines trained to: (a) reproduce standard, and (b) obtain new imaging capabilities. In the second part, we assess a NIP optimized for low-light imaging~\cite{Chen2018}. 

\subsubsection{Standard Pipeline Reproduction} We analyze NIPs trained to faithfully reproduce our standard Python pipeline (Section~\ref{sec:nip-training}). We developed all 150 images for all pipelines and cameras. We randomly divided the images into three sets: for PRNU estimation (90 images), correlation predictor training (30), and evaluation (30). We used the same split for all pipelines to ensure that the same images are used at every stage. To speed-up processing, we operate on 1~Mpx central crops. For our experiments, we used the PRNU toolbox from~\cite{Korus2016TIFS,gihub-multiscale-prnu}. Camera fingerprints were obtained using the standard MLE estimator with wavelet-based denoising~\cite{dde}. 

We report results for Nikon D90 (we observed similar results for other cameras; see supplement), for which we additionally repeated \emph{UNet} training 5 times to assess pipeline stability. Fig.~\ref{fig:prnu}a shows cross-pipeline correlations between camera fingerprints. Outside the diagonal, the values range from 0.57 to 0.96 and cluster in 3 groups (the outlying 0.57, a two large cluster around 0.75 and 0.9). Overall, \emph{libRAW} and the 4th instance of \emph{UNet} revealed the weakest similarity with the remaining pipelines. Visual inspection reveals no differences between the output of different \emph{UNet} instances. 

To measure the impact of pipeline mismatch on actual forensics tasks, we performed a cross-pipeline tampering localization experiment. We generated synthetic tampered photographs (for all pipelines), and used the standard tampering localization protocol with camera models (PRNU fingerprint and correlation predictors) obtained from every pipeline. We set analysis window size to $129 \times 129$~px, stride to 8~px, and used central-pixel attribution~\cite{Korus2016TIFS}. The tampering involved horizontally flipping a central $256\times 256$~px region. To assess tampering localization performance, we measured the average of best $F_1$ scores (over decision thresholds) for 30 images per pipeline/pipeline configuration. 

The results (Fig.~\ref{fig:prnu}b) demonstrate that some of the pipelines deliver visibly worse performance - in particular the 2nd and 4th instances of \emph{UNet}. The $F_1$ scores range between 0.35 and 0.51, which disappoints given favorable setup of the experiment (same images used at every stage, no compression, no post-processing). The qualitative impact on tampering localization is shown in Fig.~\ref{fig:prnu}c which illustrates the obtained tampering probability maps for an example test image. It can be observed that pipeline mismatch can lead to visibly worse localization results - ranging from increased background noise up to completely unreliable results. 

\subsubsection{Novel Imaging Capabilities}

\begin{table}[t]
	\centering
	\caption{Cross-pipeline correlation of PRNU fingerprints for a Sony ILCE-7SM2 camera (short-long exposure mapping).}
	\label{tab:cross-pipeline-prnu}
    \resizebox{1.0\columnwidth}{!}{
	\begin{tabular}{lcccr}
	\toprule 
        & \textbf{libRAW} & \textbf{libRAW} & \textbf{UNet} & \textbf{Average} \tabularnewline
        & \textbf{(long)} & \textbf{(short)} & \textbf{(short)} & \textbf{PCE} \tabularnewline
	\midrule
	\textbf{libRAW (long-exposure)}  & 1.00 & 0.06 & 0.08 & 470 \tabularnewline
	\textbf{libRAW (short-exposure)} & -    & 1.00 & 0.09 & 807 \tabularnewline
	\textbf{UNet (short-exposure)}   & -    & -    & 1.00 &  13 \tabularnewline
	\bottomrule
    \end{tabular}
    }
\end{table}

We focus on a NIP trained to develop photographs captured in low-light conditions~\cite{Chen2018}. We used the pre-trained UNet model available online~\cite{Chen2018github}. The experiment included two pipelines: \emph{libRAW} used on short-exposure, and long-exposure photographs, and the UNet model trained to map from short to long exposures. Analogously to the previous experiment, we crop a 1~Mpx square from the image center. We used 100 randomly selected photographs for PRNU estimation, and 20 images for measuring the fingerprint match using peak-to-correlation energy (PCE). The results (Tab.~\ref{tab:cross-pipeline-prnu}) show no correlation between any pairs of pipelines. While using \emph{libRAW} allowed for successful PRNU-based source attribution (high PCE scores), the UNet model has rendered PRNU fingerprinting ineffective. We were surprised to see a complete lack of correlation between the two libRAW-developed configurations. We attribute this to heavier post-processing needed to develop reasonable images from short-exposure photographs, and stronger contamination of the fingerprint estimate by noise. 

\subsubsection{Conclusions} Our experiments indicate that adoption of complex computational models can impact existing forensic protocols, even when they rely on traces of seemingly unaltered components. Depending on the processing history, analysts should take into account the imaging pipeline used to produce the photographs. A performance penalty can be expected when using different pipelines for building/using the models - e.g., native JPEGs produced by the camera vs professional dark room software (and its different vendors). Recent research in steganography argues that photographs can exhibit different statistics, and may need to be considered as originating from different sources when acquisition settings change~\cite{bas2016steganography,gibouloto2018steganalysis}. The issue deserves a separate rigorous study, and is an interesting direction for future work.

\section{End-to-end Optimization of Imaging and Distribution Pipelines}
\label{sec:pipeline-optimization}

\begin{figure*}[!t]
    \includegraphics[width=1.0\textwidth]{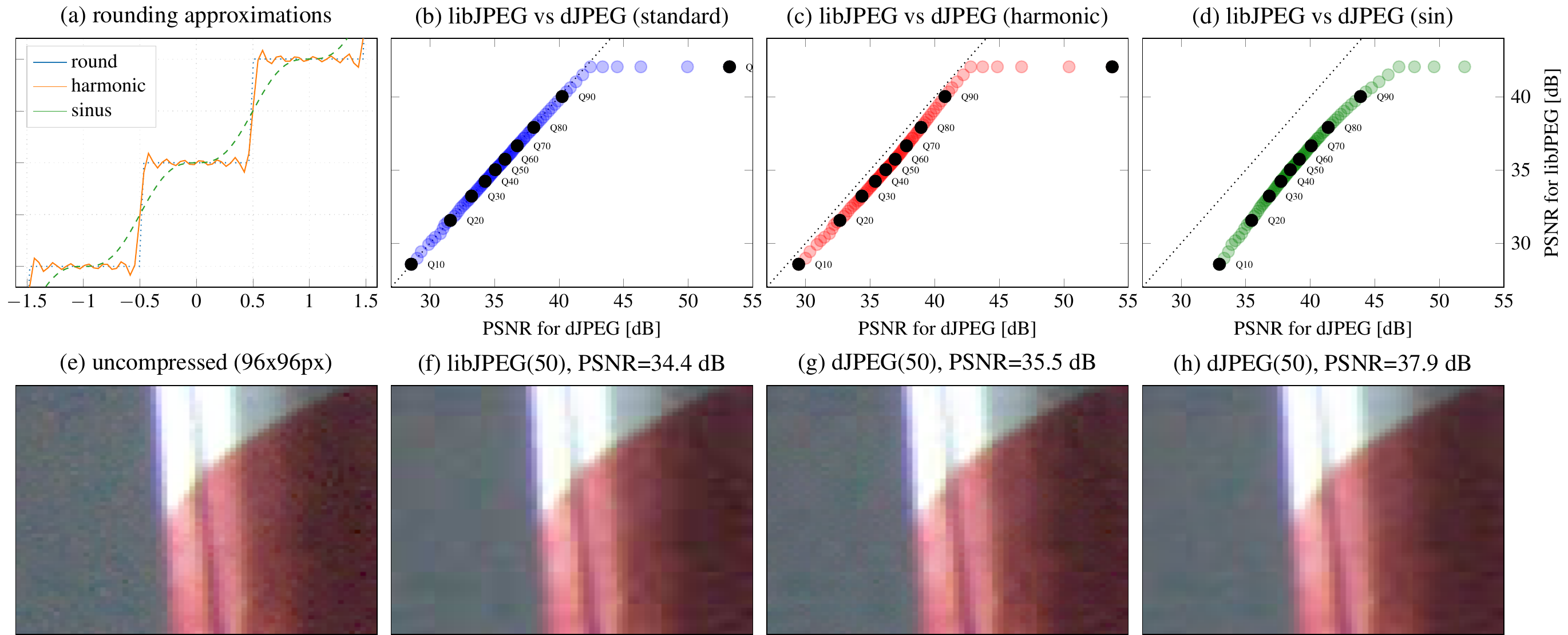}
    \vspace{0.0cm}
    \caption{Implementation of JPEG compression as a fully differentiable \emph{dJPEG} module: (a) continuous approximations of the rounding function; (b)-(d) validation of the dJPEG module against the standard \emph{libJPEG} library with standard rounding, and the harmonic and sinusoidal approximations; (e) an example image patch; (f) standard JPEG compression with quality 50; (g)-(h) dJPEG-compressed patches with the harmonic and sinusoidal approximations.}
    \label{fig:jpegnet}
\end{figure*}

One of the main limitations of conventional forensics is its reliance on fragile statistical traces. Down-sampling and compression employed by media distribution channels destroy most of the relevant traces and inhibit forensic analysis.

The core of the proposed approach is to model the entire acquisition and distribution channel, and optimize the neural imaging pipeline (NIP) to facilitate photo provenance analysis after content distribution (Fig.~\ref{fig:nip-training-protocol}). The analysis is performed by a forensic analysis network (FAN) which makes decisions about the authenticity/processing history of the analyzed photographs, and can provide feedback to the NIP. In the presented example, the model is trained to perform manipulation detection, i.e., to classify images as either straight from the camera, or as affected by a certain post-processing operation. The distribution channel mimics the behavior of modern photo sharing services and social networks which habitually down-sample and re-compress photographs. As will be demonstrated later, forensic analysis in such conditions is unreliable.

The parameters of the NIP are updated to guarantee both faithful representation of target color photographs ($L_2$ loss), and accurate decisions in post-distribution forensic analysis (cross-entropy loss). Hence, the minimized loss function $L$ can be expressed as:
\begin{subequations}
    \begin{align}    
    L &= L_{\text{fan}} + \lambda L_{\text{nip}} \\
    L_{\text{fan}} &= \frac{1}{N} \sum_{n} \sum_{c} \text{log} \Big( \text{fan}_{c} (d_c( \text{nip}(x_n~|~\theta_{\text{nip}})  )~|~\theta_{\text{fan}}) \Big) \\
    L_{\text{nip}} &= \frac{1}{3whN} \sum_{n} 255^2 \| y_n - \text{nip}(x_n~|~\theta_{\text{nip}}) \|_2 
\end{align}
\end{subequations}
\noindent where: $\theta_{\text{nip/fan}}$ are the parameters of the NIP and FAN networks, respectively; $x_n$ are the raw sensor measurements for the $n$-th example patch; $y_n$ is the corresponding target color image; $\text{nip}(x_n)$ is the image developed by the NIP from $x_n$; $d_c()$ denotes an image processed by manipulation $c$; $\text{fan}_c()$ is the probability that an image belongs to the $c$-th manipulation class, as estimated by the FAN model. Computation of the $L_2$ loss with signals in the [0,255] range makes the loss components more easily comparable, and simplifies the choice of regularization strength $\lambda$. By changing $\lambda$, we can control the trade-off between the image fidelity and forensics' accuracy.

\subsection{Modeling the Distribution Channel}

In our experiment, the RGB image developed by the NIP can undergo one of 4 possible manipulations before it is sent through a distribution channel (see Sec.~\ref{sec:manipulation} for manipulation details). The channel is modeled as subsequent down-sampling by factor 1:2 and JPEG compression - a combination that is habitually used by online photo sharing services. To enable end-to-end optimization of the entire acquisition and distribution channel, we need to ensure that every processing step remains differentiable. The main problem is JPEG compression, which requires approximation. Our channel uses quality level 50, which PSNR-wise corresponds to approx. quality 80 of the standard codec (see Fig.~\ref{fig:jpegnet}d). 

\subsection{Approximation of JPEG Compression}
\label{sec:jpeg}

We designed a \emph{dJPEG} model which approximates the standard JPEG codec, and expresses its successive steps as matrix multiplications or convolution layers that can be implemented in TensorFlow (see supplementary materials for a detailed network architecture):
\begin{itemize}
    \itemsep0em 
    \item RGB to/from YCbCr color-space conversions are implemented as $1\times1$ convolutions.
    \item Isolation of $8\times8$ blocks for independent processing is implemented by combining \emph{space-to-depth} and reshaping operations.
    \item 2D discrete cosine transforms are implemented by matrix multiplication ($DxD^{T}$ where $x$ denotes an $8\times8$ array, and $D$ is the transformation matrix).
    \item Division/multiplication of DCT coefficients by the corresponding quantization steps are implemented as element-wise operations with tiled and concatenated quantization matrices (both the luminance and chrominance channels).
    \item The actual quantization is approximated by a continuous function $\rho(x)$ - see details below.
\end{itemize}

The key problem in making JPEG fully differentiable lies in the rounding of DCT coefficients. Initially, we experimented with a Taylor series expansion, which can be made arbitrarily accurate by including more terms. Finally, we decided to use a smoother, and simpler sinusoidal approximation obtained by matching the phase of a sinusoid with the sawtooth function:
\begin{equation}
    \rho(x) = x -  \frac{\text{sin}(2 \pi x)}{2\pi}
\end{equation}
\noindent Both approximations are shown in Fig.~\ref{fig:jpegnet}a. 

The \emph{dJPEG} model relies on standard quantization matrices, derived from a desired quality level. We validated our model by comparing output images with a reference codec from \emph{libJPEG}. The results are shown in Fig.~\ref{fig:jpegnet}bcd for standard rounding, and the two approximations, respectively. We used 5 terms for the harmonic approximation. The developed module produces equivalent compression results with standard rounding, and a good approximation for its differentiable variants. Fig.~\ref{fig:jpegnet}e-h show a visual comparison of an example image patch, and its \emph{libJPEG} and \emph{dJPEG}-compressed counterparts.

\subsection{Image Manipulation}
\label{sec:manipulation}

\begin{figure}[t]
    \centering
    \includegraphics[width=1.0\columnwidth]{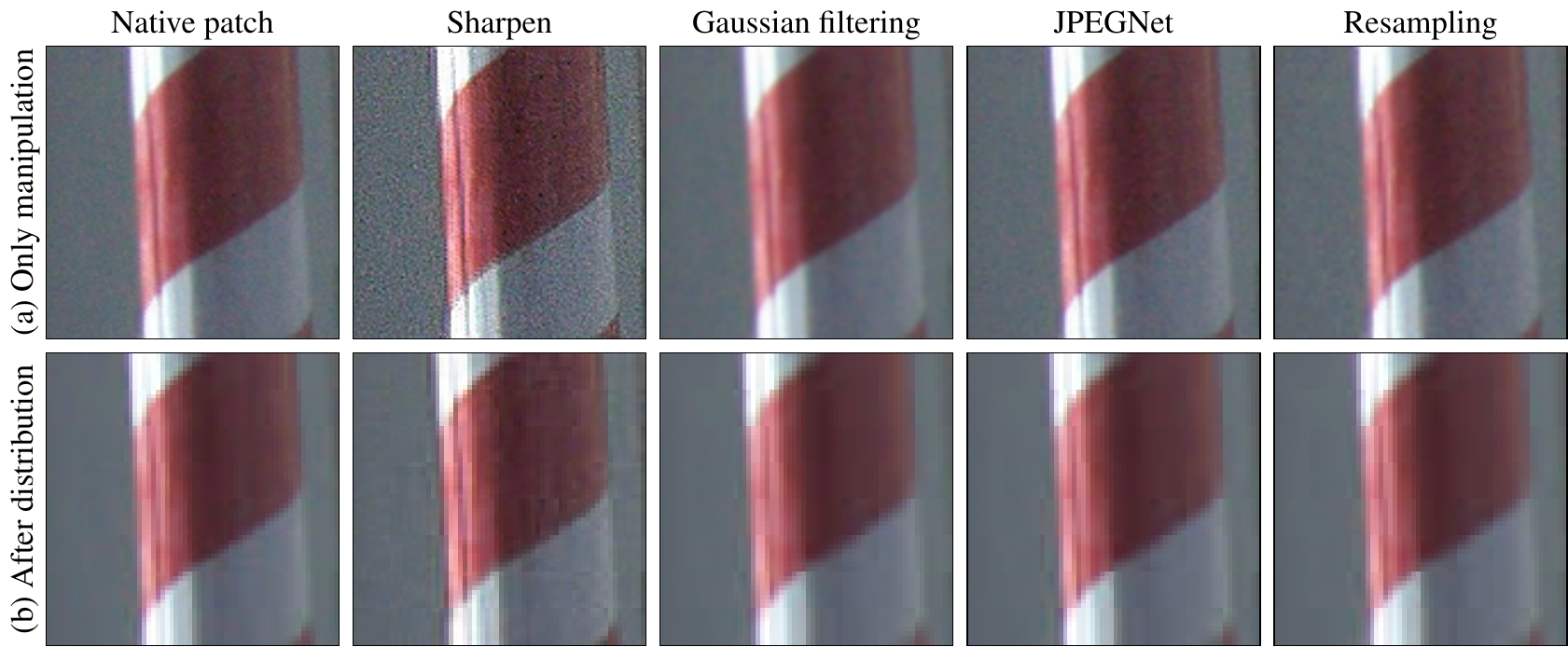}
    \caption{An example image patch with all of the considered manipulation variants: (a) just after the manipulation; (b) after the distribution channel (down-sampling and JPEG compression).}
    \label{fig:post-processing}
\end{figure}

Our experiment mirrors the standard setup for image manipulation detection~\cite{Fan2015,bayar2018constrained,boroumand2018deep}. We consider four mild post-processing operations, which may be used to mask the traces of a prospective forgery: 

\begin{itemize}
    \itemsep0em 
    \item \emph{sharpening} - implemented as an unsharp mask operator with the following kernel: 
    \begin{eqnarray}
        \begin{footnotesize}
        \frac{1}{6}
        \begin{bmatrix}
            -1 & -4 & -1 \\
            -4 & 26 & -4 \\
            -1 & -4 & -1 \\
        \end{bmatrix}
    \end{footnotesize}
    \end{eqnarray}
    applied to the luminance channel in the HSV color space.
    \item \emph{resampling} - implemented as successive down-sampling and up-sampling using bilinear interpolation and scaling factors 1:2 and 2:1.
    \item \emph{Gaussian filtering} - implemented using a convolutional layer with a $5\times5$ filter and standard deviation 0.83.
    \item \emph{JPG compression} - implemented using dJPEG with sinusoidal rounding approximation and quality level 80.
\end{itemize}

Fig.~\ref{fig:post-processing} shows post-processed variants of an example image patch: (a) just after manipulation; and (b) after the distribution channel (as seen by the FAN model). 

\subsection{The Forensic Analysis Network}

The forensic analysis network (FAN) is implemented as a CNN following the most recent recommendations on construction of neural networks for forensics analysis~\cite{bayar2018constrained}. Bayar and Stamm have proposed a new layer type, which constrains the learned filters to be valid residual filters~\cite{bayar2018constrained}. Adoption of the layer helps ignore visual content and facilitates extraction of forensically-relevant low-level features. In summary, our network operates on $128 \times 128 \times 3$ patches in the RGB color space and includes (see supplementary materials for a complete network architecture):
\begin{itemize}
    \itemsep0em 
    \item A constrained convolutions layer learning $5 \times 5$ residual filters and with no activation function.
    \item Four $5 \times 5$ convolutional layers with doubling number of feature maps (starting from 32). The layers use leaky ReLU activation and are followed by $2\times2$ max pooling.
    \item A $1\times1$ convolutional layer mapping 256 features into 256 features for each spatial location.
    \item A global average pooling layer reducing the total number of features to 256.
    \item Two fully connected layers with 512 and 128 nodes activated by leaky ReLU.
    \item A fully connected layer with $C=5$ output nodes and softmax activation.
\end{itemize}

\noindent In total, the network has 1,341,990 parameters. The network yields as an output the probabilities that characterize the processing history of each patch (4 considered manipulation classes + unprocessed/straight from the camera).

\section{Experimental Evaluation}

We start our evaluation by validating the implemented FAN model. Initially, we use the network to detect image manipulations (Sec.~\ref{sec:manipulation}) without a distribution channel (Sec.~\ref{sec:fan-validation}). Then, we perform extensive evaluation of the entire acquisition and distribution workflow (Sec.~\ref{sec:secure-nip-results}) and demonstrate the benefits of jointly optimizing the FAN and NIP models.

\subsection{FAN Model Validation}
\label{sec:fan-validation}

To validate our FAN model, we used it to directly discriminate between the manipulated patches (no distribution channel distortion, as in~\cite{bayar2018constrained}). We used the \emph{UNet} model as the imaging pipeline, and adjusted the raw input size to guarantee same-size inputs to the FAN ($128\times128\times3$ RGB images). In such conditions, the model yields classification accuracy of 99\%, which is consistent with~\cite{bayar2018constrained}.

\subsection{Imaging Pipeline Optimization}
\label{sec:secure-nip-results}

\begin{figure}
    \centering
    \includegraphics[width=1.00\columnwidth]{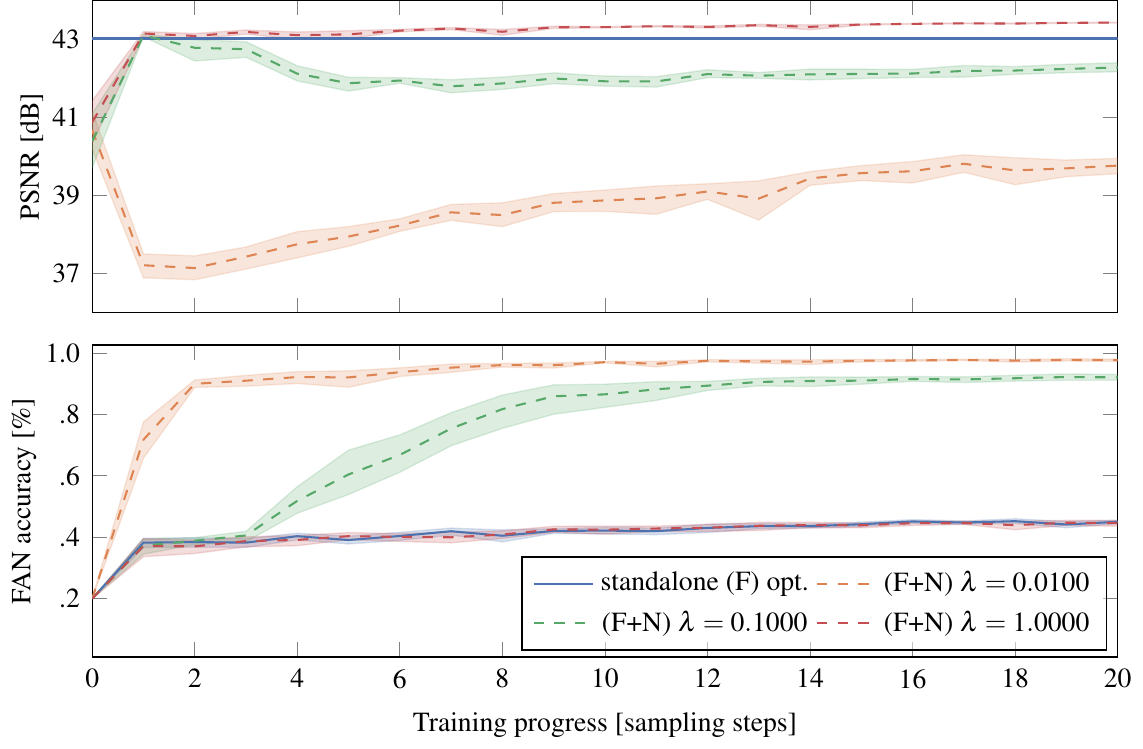}
    \caption{Typical progression of validation metrics (Nikon D90) for standalone FAN training (F) and joint optimization of FAN and NIP models (F+N).}
    \label{fig:training-progress}
\end{figure}

\begin{figure*}
    \centering
    \includegraphics[width=1.00\textwidth]{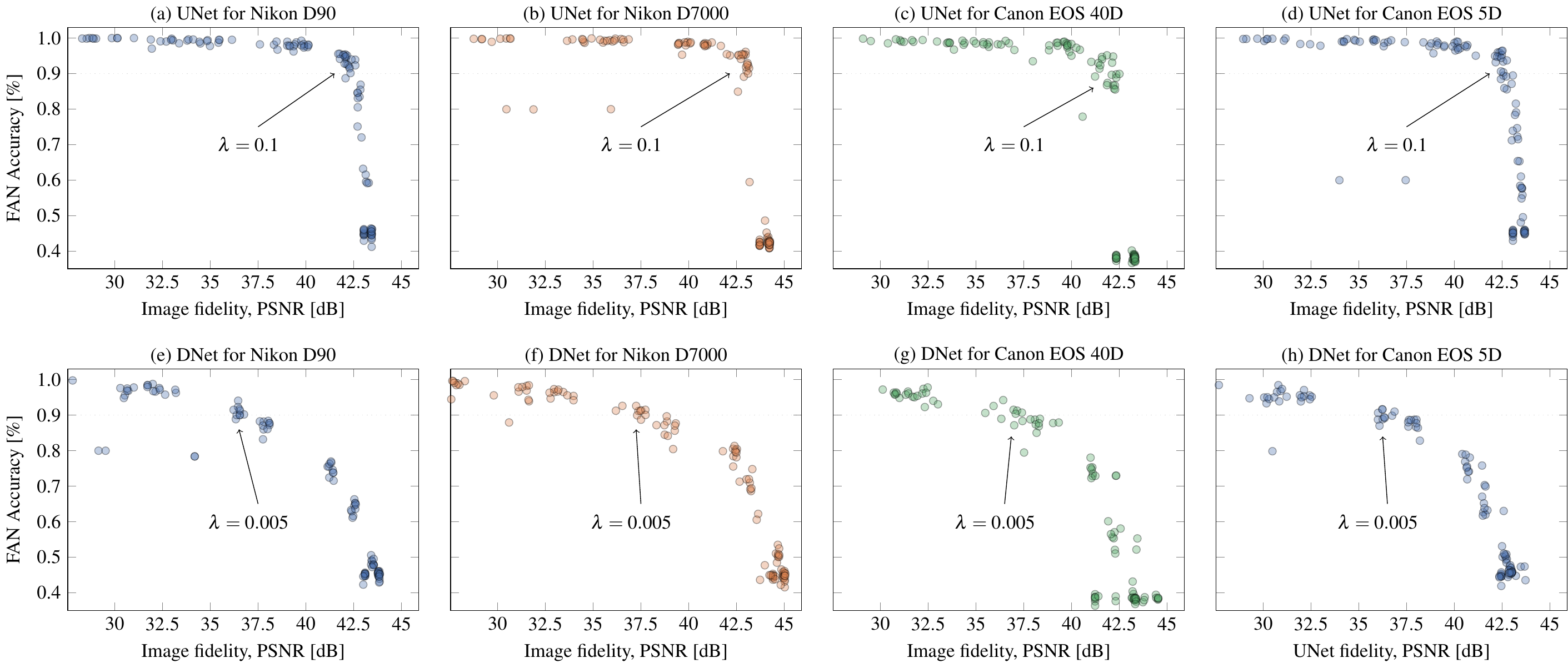}
    \caption{Trade-off between the photo development fidelity (in PSNR [dB]) and the forensic decision accuracy (FAN accuracy at the end of the distribution channel) for regularization strength $\lambda$ between $10^{-4}$ and $10^{0}$. Each configuration was trained 10 times with different random parameter initialization.}
    \label{fig:scatterplots}
\end{figure*}

Our main experiment addresses discrimination of manipulation classes at the end of a distribution channel where down-sampling and lossy compression make it more challenging to tell the manipulations apart (Fig.~\ref{fig:post-processing}). We consider two optimization modes: (F) only the FAN network is optimized given a fixed NIP model; (F+N) both the FAN and NIP models are optimized jointly. In the latter mode, we use the regularization strength $\lambda$ to control the trade-off between image fidelity and classification accuracy. In both cases, the NIPs are pre-initialized with previously trained models (Section~\ref{sec:pipelines}).

Similarly to previous experiments, we used 120 images for training, and the remaining 30 images for validation. In each iteration we randomly extract new patches from full-resolution training images. The validation set is fixed at the beginning and includes 100 random patches per image (3,000 in total) for classification accuracy assessment. To speed-up evaluation, we used 2 random patches per image (60 in total) for image quality assessment\footnote{We measured both the PSNR and SSIM metrics. However, we omit SSIM results since they show the same qualitative behavior as PSNR. Moreover, due to a wider range of metric values, the discussed effects are better visible in PSNR measurements.}. To prevent over-representation of empty content, we reject patches with pixel variance $<$ 0.01, and keep the ones with variance $<$ 0.02 with 50\% chance. More diverse patches are always accepted. 

We performed the experiment for 4 cameras (Canon EOS 40D and EOS 5D, and Nikon D7000, and D90 - see Tab.~\ref{tab:cameras}). To assess optimization trade-offs, we tested 11 values of the regularization strength $\lambda$ ranging from $10^{-4}$ up to $1$. Each run was repeated 10 times with random initialization of FAN's parameters. Due to computational constraints, we ran the optimization for 1,000 epochs, starting with a learning rate of $10^{-4}$ and systematically decreasing by 15\% every 100 epochs. This was enough for consistent convergence of \emph{INet} and \emph{UNet} models (Fig.~\ref{fig:training-progress}). For some regularization strengths $\lambda$, the \emph{DNet} model could potentially benefit from longer training.

In the standalone (F) optimization mode, the FAN delivers only $\approx$ 45\% accuracy after the distribution channel (see Tab.~\ref{tab:confusion}a for a typical confusion matrix). This result is stable across all cameras and pipelines. In the (F+M) mode, we observed consistent improvements. Due to limited expressive power, the \emph{INet} model delivered marginal gains in accuracy - up to $\approx$ 55\% with a reasonable image distortion. The \emph{DNet} and \emph{UNet} models allowed to boost manipulation detection accuracy to $\approx$ 75\% and $\approx$ 90\%, respectively. Given more permissive quality constraints, both models allow for nearly perfect manipulation detection. Hence, we focus on these two models.

A typical progression of validation metrics (classification accuracy and distortion PSNR) for the \emph{UNet} model is shown in Fig.~\ref{fig:training-progress} (sampled every 50 epochs). In the (F) optimization mode, classification accuracy saturates at $\approx$ 45\%. Similar effect can be observed for joint (F+N) optimization with strong regularization ($\lambda=1$), which saturates at the same level, but leads to slow improvements in image quality (which indicates that the pre-trained NIP models can be slightly improved given longer training time). Intermediate regularization strengths allow to trade image fidelity for reliable manipulation detection. For $\lambda=10^{-1}$ and $10^{-2}$ the classification accuracy saturates at 92\% and 98\%, respectively. The \emph{DNet} model exhibited similar behavior, but converged slower and eventually yielded a worse quality-performance trade-off (Fig.~\ref{fig:scatterplots}).

To facilitate forensic analysis, the joint (F+N) optimization mode learns to introduce carefully crafted distortions at the time of photo development. Fig.~\ref{fig:scatterplots} shows the quality-accuracy trade-offs for all of the evaluated models. The obtained curves are promising, and suggest that only minor quality degradation may be sufficient to obtain significant gains in manipulation detection performance. \emph{UNet} improved rapidly for all camera models, and reached classification accuracy $\approx$ 90\% while maintaining average image representation fidelity of $\approx$ 42.5~dB ($\lambda=0.1$). Typical confusion matrices for various regularization strengths are shown in Tab.~\ref{tab:confusion}. \emph{DNet} delivered worse performance, and visually acceptable distortion resulted in forensics accuracy of $\approx$ 75\%. We acknowledge that actual visibility of the introduced artifacts depends on image content, and tends to be masked in textured areas. 

\begin{table}[t]
    \caption{Typical confusion matrices (Nikon D90). Entries $\approx{}0$ are not shown; entries $<3\%$ are marked with (*).}
    \label{tab:confusion}
    \vspace{4pt}
    \centering
    \resizebox{0.85\columnwidth}{!}{
    \begin{tabular}{lrrrrr}
        \multicolumn{6}{c}{(a) standalone FAN opt. (UNet) $\rightarrow$ 44.6\%} \tabularnewline   
        \diagbox{\textbf{True}}{\textbf{Predicted}} & \rotatebox{90}{\textbf{nat.~}}  &  \rotatebox{90}{\textbf{sha.}}  &  \rotatebox{90}{\textbf{gau.}}  &  \rotatebox{90}{\textbf{jpg}}  &  \rotatebox{90}{\textbf{res.}}  \tabularnewline
        \toprule
        \textbf{native} & \cellcolor{lime!36} 36& \cellcolor{red!17} 17& \cellcolor{red!14} 14& \cellcolor{red!23} 23& \cellcolor{red!10} 10 \tabularnewline
        \textbf{sharpen} & \cellcolor{red!5} 5& \cellcolor{lime!91} 91& *& *& * \tabularnewline
        \textbf{gaussian} & \cellcolor{red!11} 11& \cellcolor{red!3} 3& \cellcolor{lime!50} 50& \cellcolor{red!15} 15& \cellcolor{red!21} 21 \tabularnewline
        \textbf{jpg} & \cellcolor{red!34} 34& \cellcolor{red!16} 16& \cellcolor{red!14} 14& \cellcolor{lime!25} 25& \cellcolor{red!11} 11 \tabularnewline
        \textbf{resample} & \cellcolor{red!19} 19& \cellcolor{red!6} 6& \cellcolor{red!30} 30& \cellcolor{red!25} 25& \cellcolor{lime!21} 21 \tabularnewline
        \bottomrule
        \tabularnewline

        \multicolumn{6}{c}{(b) joint FAN+NIP opt. (UNet) $\lambda=0.25$  $\rightarrow$ 87.0\% }\tabularnewline        
        \diagbox{\textbf{True}}{\textbf{Predicted}} & \rotatebox{90}{\textbf{nat.~}}  &  \rotatebox{90}{\textbf{sha.}}  &  \rotatebox{90}{\textbf{gau.}}  &  \rotatebox{90}{\textbf{jpg}}  &  \rotatebox{90}{\textbf{res.}}  \tabularnewline
        \toprule
        \textbf{native} & \cellcolor{lime!77} 77& *& *& \cellcolor{red!20} 20&  \tabularnewline
        \textbf{sharpen} & *& \cellcolor{lime!98} 98& & &  \tabularnewline
        \textbf{gaussian} & *& & \cellcolor{lime!93} 93& \cellcolor{red!6} 6&  \tabularnewline
        \textbf{jpg} & \cellcolor{red!23} 23& *& \cellcolor{red!3} 3& \cellcolor{lime!73} 73&  \tabularnewline
        \textbf{resample} & *& & *& \cellcolor{red!3} 3& \cellcolor{lime!94} 94 \tabularnewline
        
        \bottomrule
        \tabularnewline

        \multicolumn{6}{c}{(c) joint FAN+NIP opt. (UNet) $\lambda=0.1$ $\rightarrow$ 92.6\%} \tabularnewline
        \diagbox{\textbf{True}}{\textbf{Predicted}} & \rotatebox{90}{\textbf{nat.~}}  &  \rotatebox{90}{\textbf{sha.}}  &  \rotatebox{90}{\textbf{gau.}}  &  \rotatebox{90}{\textbf{jpg}}  &  \rotatebox{90}{\textbf{res.}}  \tabularnewline
        \toprule
        \textbf{native} & \cellcolor{lime!85} 85& & *& \cellcolor{red!13} 13&  \tabularnewline
        \textbf{sharpen} & *& \cellcolor{lime!97} 97& & *&  \tabularnewline
        \textbf{gaussian} & \cellcolor{red!6} 6& & \cellcolor{lime!93} 93& &  \tabularnewline
        \textbf{jpg} & \cellcolor{red!6} 6& *& & \cellcolor{lime!91} 91&  \tabularnewline
        \textbf{resample} & & & \cellcolor{red!3} 3& & \cellcolor{lime!97} 97 \tabularnewline    
        \bottomrule       
        \tabularnewline

        \multicolumn{6}{c}{(d) joint FAN+NIP opt. (UNet) $\lambda=0.05$ $\rightarrow$ 95.8\%} \tabularnewline
        \diagbox{\textbf{True}}{\textbf{Predicted}} & \rotatebox{90}{\textbf{nat.~}}  &  \rotatebox{90}{\textbf{sha.}}  &  \rotatebox{90}{\textbf{gau.}}  &  \rotatebox{90}{\textbf{jpg}}  &  \rotatebox{90}{\textbf{res.}}  \tabularnewline
        \toprule
        \textbf{native} & \cellcolor{lime!94} 94& & *& \cellcolor{red!4} 4&  \tabularnewline
        \textbf{sharpen} & *& \cellcolor{lime!97} 97& & *& * \tabularnewline
        \textbf{gaussian} & *& & \cellcolor{lime!98} 98& & * \tabularnewline
        \textbf{jpg} & \cellcolor{red!8} 8& *& & \cellcolor{lime!90} 90&  \tabularnewline
        \textbf{resample} & & & & & \cellcolor{lime!100} 100 \tabularnewline        
        \bottomrule  
    \end{tabular}
    }
\end{table}

\begin{figure*}
    \centering
    \includegraphics[width=0.89\textwidth]{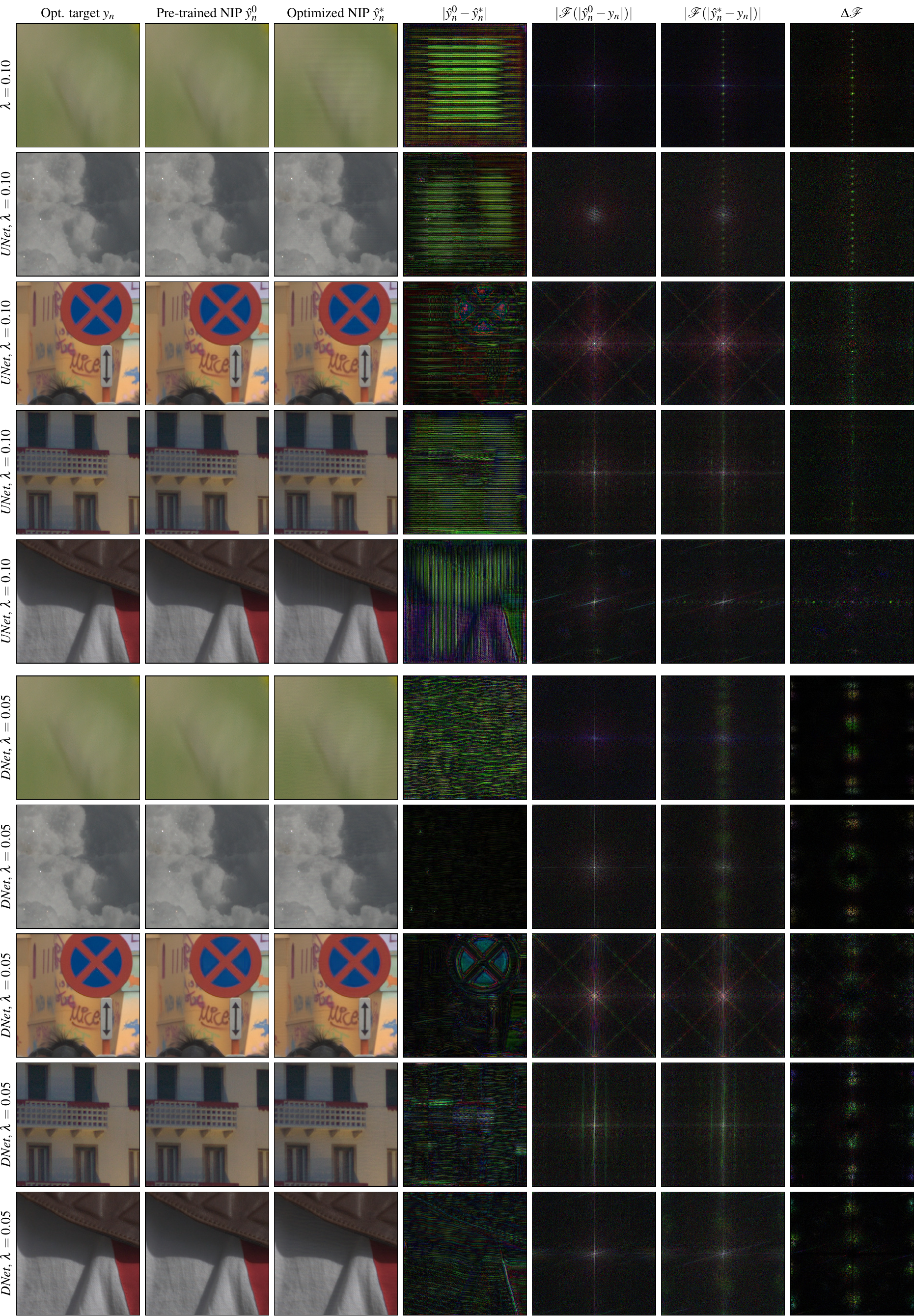}
    \caption{Example image patches ($128\times128$~px) developed by a standard NIP $\hat{y}^0_{n}$ and by a NIP jointly optimized with a FAN model $\hat{y}^*_{n}$. The difference between the patches reveals periodic distortion patterns, which manifest themselves as regularly spaced modulation of the FFT frequencies (see difference between FFT amplitude spectra). The spectra are shown shown in logarithmic scale $f(x) = \text{log}(10+|x|)$.}
    \label{fig:artifacts}
\end{figure*}

Qualitative illustration of the observed artifacts is shown in Fig.~\ref{fig:artifacts}, which shows diverse image patches ($128\times128$~px) from various cameras. Corresponding results for \emph{UNet} and \emph{DNet} are shown in the top and bottom sections, respectively. For both models, we chose regularization strengths which lead to reasonable image quality (expected manipulation detection accuracy of $\approx$ 90\% and $\approx$ 75\% for \emph{UNet} and \emph{DNet}, respectively). The first 3 columns correspond to: the desired output image used as optimization target $y_n$; the output of a pre-trained NIP model $\hat{y}^0_n = \text{nip}(x_n~|~\theta^0_{\text{nip}})$; and the (distorted) output of a NIP optimized for forensic analysis $\hat{y}^*_n = \text{nip}(x_n~|~\theta^*_{\text{nip}})$. 

The considered NIP architectures lead to distinctive distortion patterns. For \emph{UNet} the artifacts assume the form of periodic horizontal (or vertical) patterns (best seen in the difference image $|\hat{y}^0_n - \hat{y}^*_n|$). Spectral analysis shows modulation of regularly spaced frequencies in the 2D FFT spectra $\mathcal{F}(|\hat{y}^*_n - y_n|)$. The spectral location of the distortions can be best observed in the difference between the spectra:
\begin{equation*}
    \Delta\mathcal{F} = | |\mathcal{F}(\hat{y}^0_n)| - |\mathcal{F}(\hat{y}^*_n)| |,
\end{equation*}
\noindent shown in the last column. (For better clarity, the spectra are presented in logarithmic intensity $f(x) = \text{log}(10+|x|)$.) \emph{DNet} introduces a more pleasant distortion resembling a wavy textured pattern. However, the model seems to be more sensitive than \emph{UNet} and for lower regularization strengths looses sharpness (this effect can already be observed in Fig.~\ref{fig:artifacts} where differences along some edges dominates over the regular distortion of interest). Spectral analysis reveals distribution of the artifacts across a wider range of frequencies, but exhibits a similar pattern - modulation of regularly spaced frequencies along either the horizontal or vertical direction.

\section{Discussion and Future Work}
\label{sec:conclusions}

Our work explores optimization of imaging pipelines with explicit provenance analysis objectives. We have demonstrated that such an approach can successfully address reliability problems with conventional forensic analysis in complex distribution channels. We performed end-to-end, joint optimization of a forensic analysis network and a neural imaging pipeline, and obtained significant improvements in photo manipulation detection, whose accuracy increased from $\approx{}45\%$ for state-of-the-art forensic to over 90\%. 

The improvement comes at the cost of image quality, as the enhanced photo analysis capabilities rely on learned artifacts in the developed photographs. Analysis of performance trade-offs indicates that minor fidelity deterioration is sufficient to get significant benefits. In this work we analyzed image patches in isolation, but in practical settings the final decision about an image should integrate intermediate predictions from multiple regions. Given all of the above, we believe that the proposed approach presents a promising line of research. 

The learned artifacts bear resemblance to digital watermarking, which also relies on carefully crafted distortions to carry copyright or authentication side-information. However, authentication watermarks need to be hand-crafted with allowed post-processing in mind, and allow only for binary decisions based on the watermark detection strength. Our approach represents the entire acquisition and distribution workflow as a single, fully-differentiable model which can be optimized end-to-end to enable more complex analysis. This opens new capabilities compared to existing technologies. 

Spectral analysis indicates interesting qualitative differences between the hand-crafted distortions in digital watermarks, and machine-learned artifacts. While watermarking systems were typically designed by designating spectral sub-bands (most often in middle frequencies) intended for information embedding, the distortions that we observed were regularly spaced across all frequencies. A rigorous comparison of the human-designed and machine-learned information hiding strategies constitutes another exciting prospect for future work.

The proposed approach aims to exploit neural imaging processors, which are an emerging trend in the design of novel digital cameras~\cite{zhang2017method,cox2017system}. We believe exploiting new computational capabilities to facilitate reliable photo authentication is an important direction of future work. Increasing adoption of computational methods in camera design bears the risk of invalidating many forensic analysis protocols. In our evaluation, we observed that adoption of neural imaging can have a major impact on PRNU analysis - one of the most reliable forensic tools to date. In the most optimistic setting where the NIP is trained to reproduce the results of a conventional pipeline, we observed non-negligible impact on the extracted sensor fingerprints. This can translate to diminished reliability of such analysis in case of pipeline mismatch. A NIP optimized for novel imaging capabilities has rendered sensor fingerprinting completely ineffective. A rigorous assessment of this phenomenon is another interesting topic for future research, especially with the increasing adoption of such technologies in contemporary devices~\cite{night-sight-pixel3}.

Overall, we believe the proposed approach opens many exciting research opportunities. In our future work, we're planning to explore techniques that could allow for better control over performance trade-offs and image distortion patterns. We will also focus on designing training protocols that would translate to good generalization to various forensics tasks, and more diverse distribution channels. Finally, it may be worth to consider expansion of end-to-end optimization capabilities to include the compression codecs used in the distribution channels.

\balance

\bibliographystyle{IEEEbib}
\bibliography{./references.bib}

\newpage

\end{document}


\begin{center}
    {\Large\textbf{Supplementary Materials for \\ \vspace{8pt} Neural Imaging Pipelines - the Scourge or Hope of Forensics?}}

    \vspace{8pt} Paweł Korus$^{1,2}$, and Nasir Memon$^{1}$ 

    New York University$^1$, AGH University of Science and Technology$^2$

    \url{http://kt.agh.edu.pl/~korus}

\end{center}

\vspace{20pt}

\section{Source Code}

Our neural imaging toolbox is available at \url{https://github.com/pkorus/neural-imaging}.

\section{Contents}

\begin{itemize}
    \itemsep2pt
    \item Impact of neural imaging pipelines on PRNU-based forensics (all cameras)
    \item Detailed validation statistics for all tested configurations of the improved pipelines
    \item Fidelity-accuracy trade-offs (both PSNR and SSIM)
    \item Detailed comparison of different variants of learned distortions - both UNet and DNet models
    \item Neural network architectures (INet, DNet, UNet, dJPEG, FAN)
\end{itemize}

\newpage

\begin{figure*}[p]
    \centering
    \includegraphics[width=0.8\textwidth]{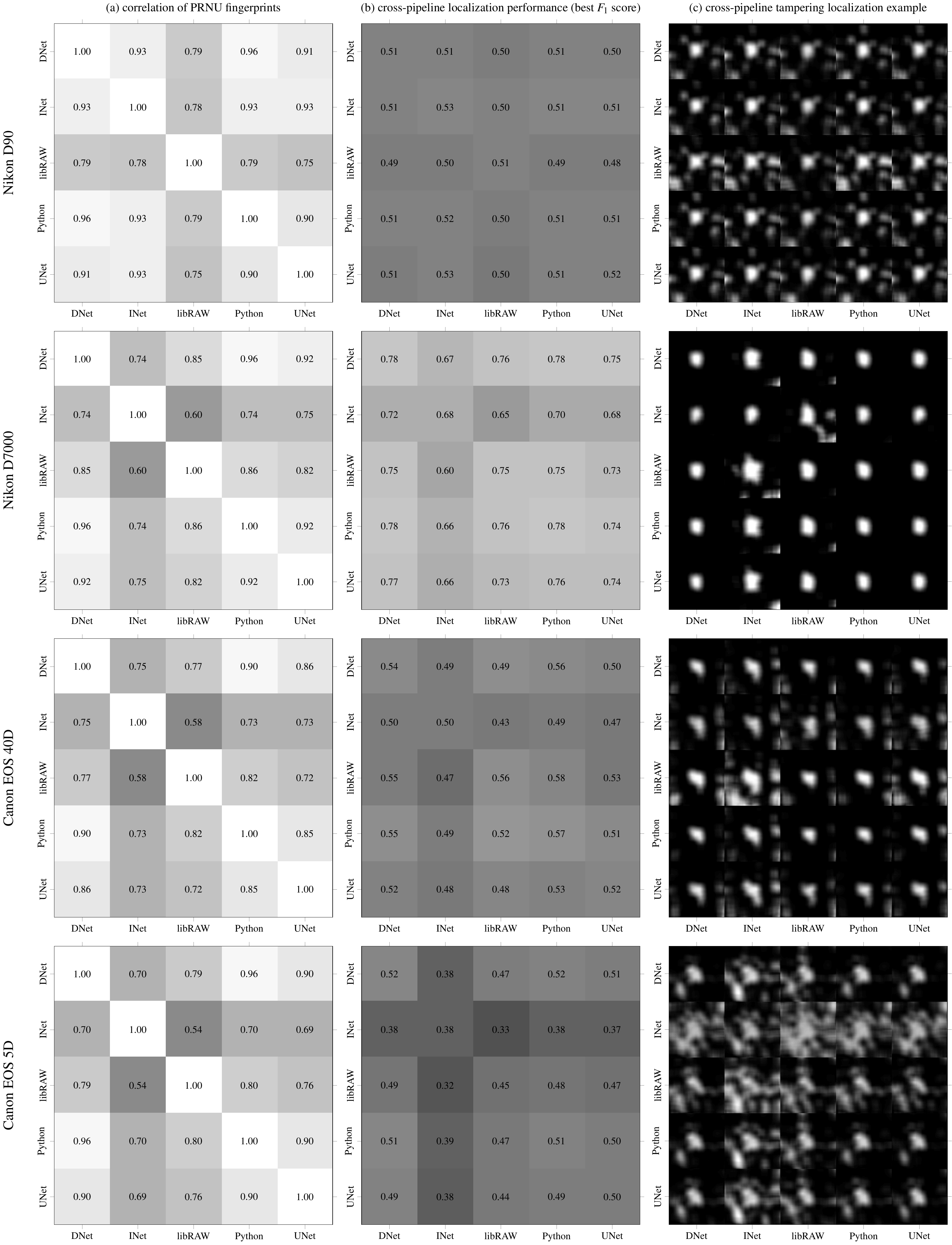}
    \caption{Impact of imaging pipeline on PRNU-based forensics analysis (a) correlation between PRNU fingerprints obtained from the same images developed by different imaging pipelines; (b) cross-pipeline tampering localization performance (best $F_1$ scores); (c) cross-pipeline tampering localization example with an attempt to localize a horizontally flipped square region in the middle (forgery size $256\times256$~px; analysis window size $129\times129$~px).}
    \label{fig:prnu-forensics-all-cameras}
\end{figure*}

\clearpage

\begin{table*}
    \caption{Validation metrics for all training configurations (averaged over 10 repetitions), }
    \label{tab:summary}
    \centering
    \begin{tabular}{llrrrrrr}
\toprule
\multirow{2}{*}{\textbf{Camera}} & \multirow{2}{*}{$\mathbf{\lambda}$} & \multicolumn{3}{c}{\textbf{DNet}} & \multicolumn{3}{c}{\textbf{UNet}} \\
\cmidrule(lr){3-5} \cmidrule(lr){6-8}
& & \textbf{PSNR [dB]} & \textbf{SSIM [0, 1]}  & \textbf{FAN acc. [\%]} & \textbf{PSNR [dB]} & \textbf{SSIM [0, 1]}  & \textbf{FAN acc. [\%]} \\
\midrule
\multirow{2}{*}{\textbf{Canon EOS 40D}} & 0.0001 & 26.7 & 0.838 & 0.964    & 30.9 & 0.900 & 0.992    \\
 & 0.0005 & 30.9 & 0.906 & 0.961    & 33.9 & 0.942 & 0.986    \\
 & 0.001  & 32.3 & 0.923 & 0.953    & 35.8 & 0.960 & 0.984    \\
 & 0.005  & 36.8 & 0.969 & 0.907    & 39.2 & 0.979 & 0.981    \\
 & 0.01   & 38.2 & 0.976 & 0.867    & 39.8 & 0.982 & 0.948    \\
 & 0.05   & 41.3 & 0.988 & 0.741    & 41.6 & 0.987 & 0.931    \\
 & 0.1    & 42.5 & 0.991 & 0.553    & 42.2 & 0.988 & 0.877    \\
 & 0.25   & 43.5 & 0.993 & 0.392    & 43.2 & 0.990 & 0.384    \\
 & 0.5    & 43.6 & 0.993 & 0.382    & 43.3 & 0.990 & 0.382    \\
 & 1      & 43.6 & 0.993 & 0.381    & 43.3 & 0.990 & 0.378    \\
 & -      & 41.5 & 0.991 & 0.383    & 42.3 & 0.989 & 0.382    \\
\midrule
\multirow{2}{*}{\textbf{Canon EOS 5D}}  & 0.0001 & 26.7 & 0.838 & 0.950    & 30.5 & 0.889 & 0.957    \\
 & 0.0005 & 30.4 & 0.897 & 0.957    & 34.6 & 0.948 & 0.951    \\
 & 0.001  & 32.0 & 0.923 & 0.951    & 35.7 & 0.961 & 0.989    \\
 & 0.005  & 36.3 & 0.961 & 0.898    & 39.4 & 0.981 & 0.984    \\
 & 0.01   & 37.9 & 0.971 & 0.873    & 40.0 & 0.984 & 0.973    \\
 & 0.05   & 40.7 & 0.982 & 0.760    & 42.2 & 0.989 & 0.953    \\
 & 0.1    & 41.7 & 0.985 & 0.649    & 42.6 & 0.990 & 0.901    \\
 & 0.25   & 42.7 & 0.987 & 0.500    & 43.2 & 0.991 & 0.779    \\
 & 0.5    & 43.0 & 0.987 & 0.464    & 43.5 & 0.991 & 0.582    \\
 & 1      & 43.0 & 0.987 & 0.454    & 43.7 & 0.991 & 0.458    \\
 & -      & 42.5 & 0.984 & 0.445    & 43.1 & 0.991 & 0.449    \\
\midrule
\multirow{2}{*}{\textbf{Nikon D7000}}   & 0.0001 & 27.8 & 0.830 & 0.986    & 30.1 & 0.891 & 0.957    \\
 & 0.0005 & 31.2 & 0.907 & 0.958    & 34.9 & 0.949 & 0.975    \\
 & 0.001  & 33.3 & 0.934 & 0.960    & 36.0 & 0.960 & 0.993    \\
 & 0.005  & 37.3 & 0.970 & 0.910    & 39.8 & 0.982 & 0.985    \\
 & 0.01   & 38.9 & 0.980 & 0.862    & 40.9 & 0.986 & 0.979    \\
 & 0.05   & 42.4 & 0.990 & 0.794    & 42.7 & 0.990 & 0.952    \\
 & 0.1    & 43.2 & 0.992 & 0.691    & 43.0 & 0.990 & 0.875    \\
 & 0.25   & 44.6 & 0.994 & 0.505    & 44.1 & 0.992 & 0.437    \\
 & 0.5    & 44.9 & 0.994 & 0.448    & 44.2 & 0.992 & 0.421    \\
 & 1      & 44.9 & 0.994 & 0.445    & 44.2 & 0.992 & 0.420    \\
 & -      & 44.4 & 0.994 & 0.444    & 43.7 & 0.992 & 0.421    \\
\midrule
\multirow{2}{*}{\textbf{Nikon D90}}     & 0.0001 & 26.6 & 0.808 & 0.938    & 29.3 & 0.865 & 0.999    \\
 & 0.0005 & 31.0 & 0.898 & 0.973    & 32.9 & 0.928 & 0.991    \\
 & 0.001  & 32.8 & 0.922 & 0.932    & 35.0 & 0.953 & 0.991    \\
 & 0.005  & 36.5 & 0.965 & 0.910    & 38.9 & 0.978 & 0.983    \\
 & 0.01   & 37.9 & 0.975 & 0.870    & 39.8 & 0.982 & 0.977    \\
 & 0.05   & 41.3 & 0.987 & 0.747    & 42.0 & 0.988 & 0.950    \\
 & 0.1    & 42.5 & 0.990 & 0.639    & 42.3 & 0.989 & 0.920    \\
 & 0.25   & 43.5 & 0.992 & 0.485    & 42.8 & 0.990 & 0.799    \\
 & 0.5    & 43.8 & 0.992 & 0.455    & 43.3 & 0.991 & 0.511    \\
 & 1      & 43.8 & 0.992 & 0.446    & 43.4 & 0.991 & 0.446    \\
 & -      & 43.1 & 0.992 & 0.448    & 43.0 & 0.990 & 0.450    \\
\bottomrule
\end{tabular}
\end{table*}

\clearpage

\begin{figure*}[p]
    \centering
    \includegraphics[width=1.0\textwidth]{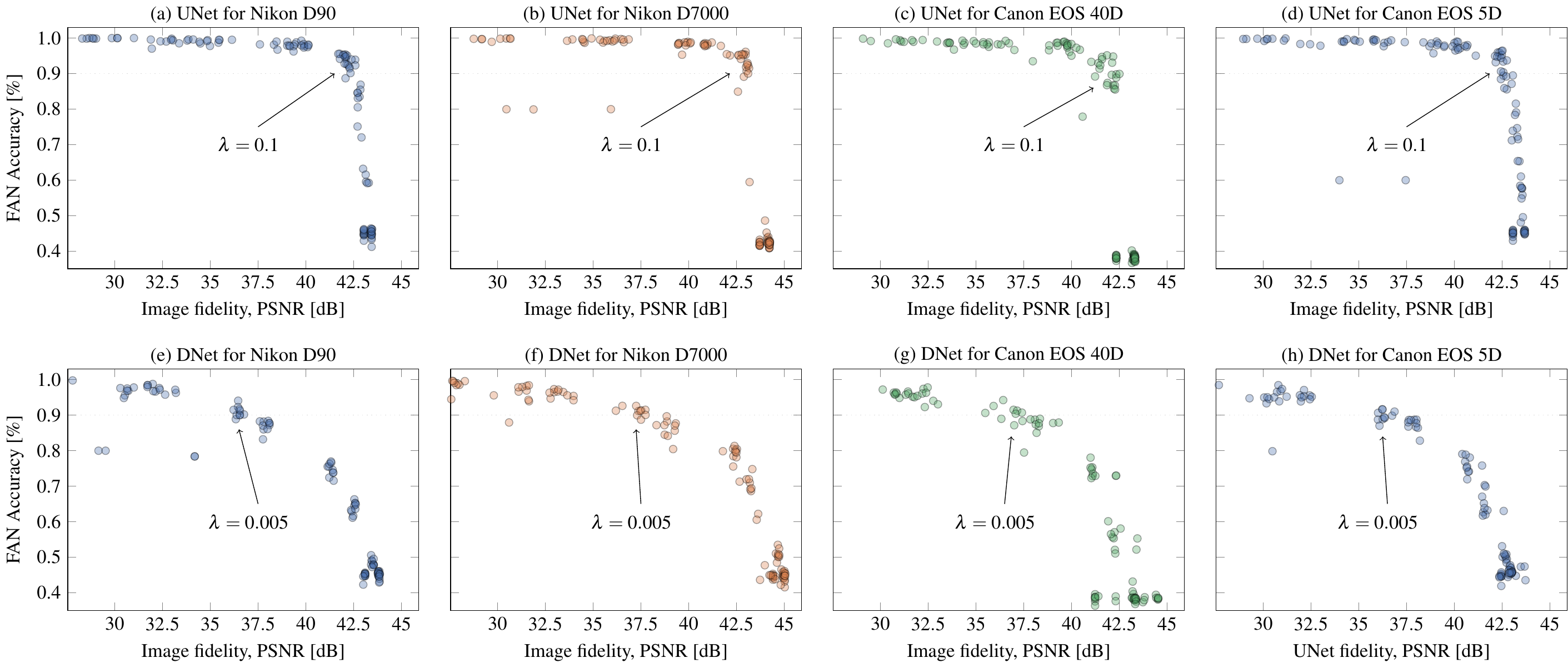}
    \caption{Trade-off between the photo development fidelity (PSNR) and the forensic decision accuracy (FAN accuracy at the end of the distribution channel) for regularization strength $\lambda$ between $10^{-4}$ and $10^{0}$. Each configuration was trained 10 times with different random parameter initialization.}
    \label{fig:scatterplots-ssim}
\end{figure*}

\begin{figure*}[p]
    \centering
    \includegraphics[width=1.0\textwidth]{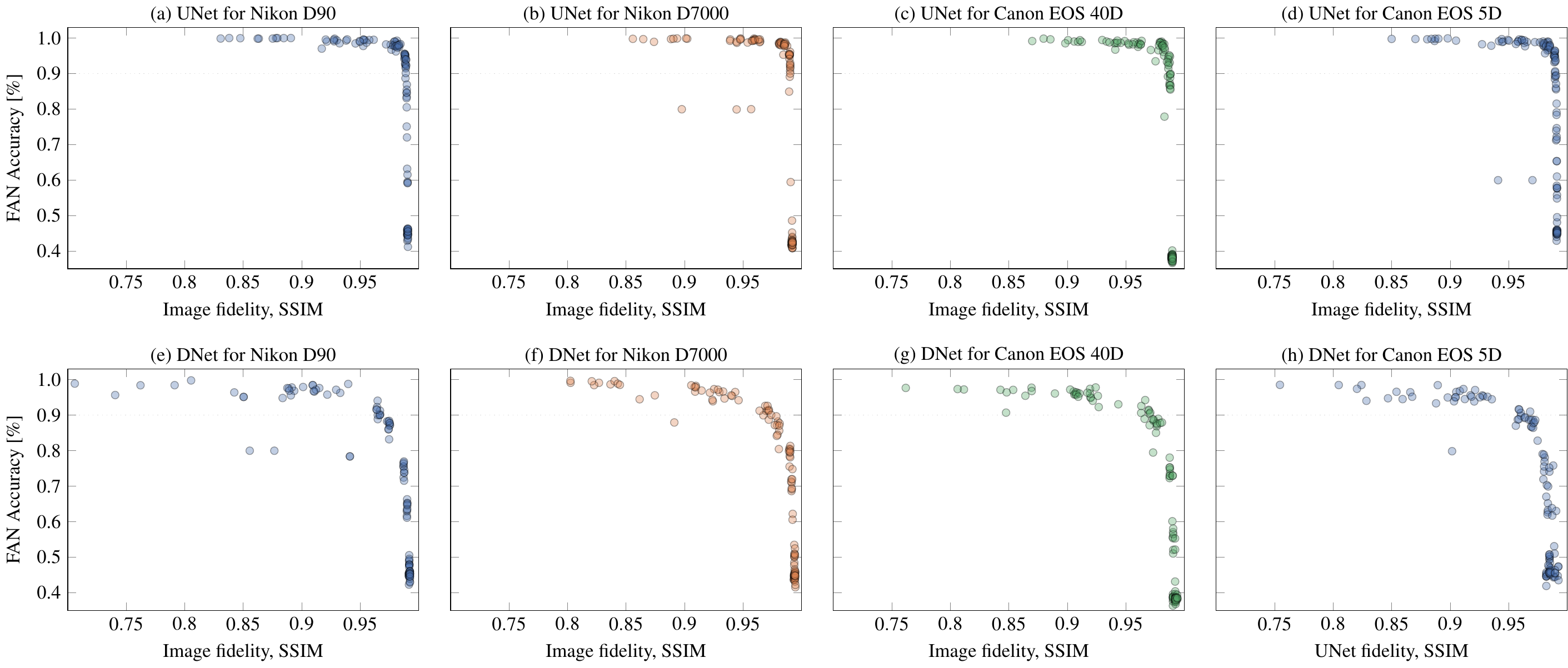}
    \caption{Trade-off between the photo development fidelity (SSIM) and the forensic decision accuracy (FAN accuracy at the end of the distribution channel) for regularization strength $\lambda$ between $10^{-4}$ and $10^{0}$. Each configuration was trained 10 times with different random parameter initialization.}
    \label{fig:scatterplots-ssim}
\end{figure*}

\begin{figure*}[p]
    \centering
    \includegraphics[width=0.90\textwidth]{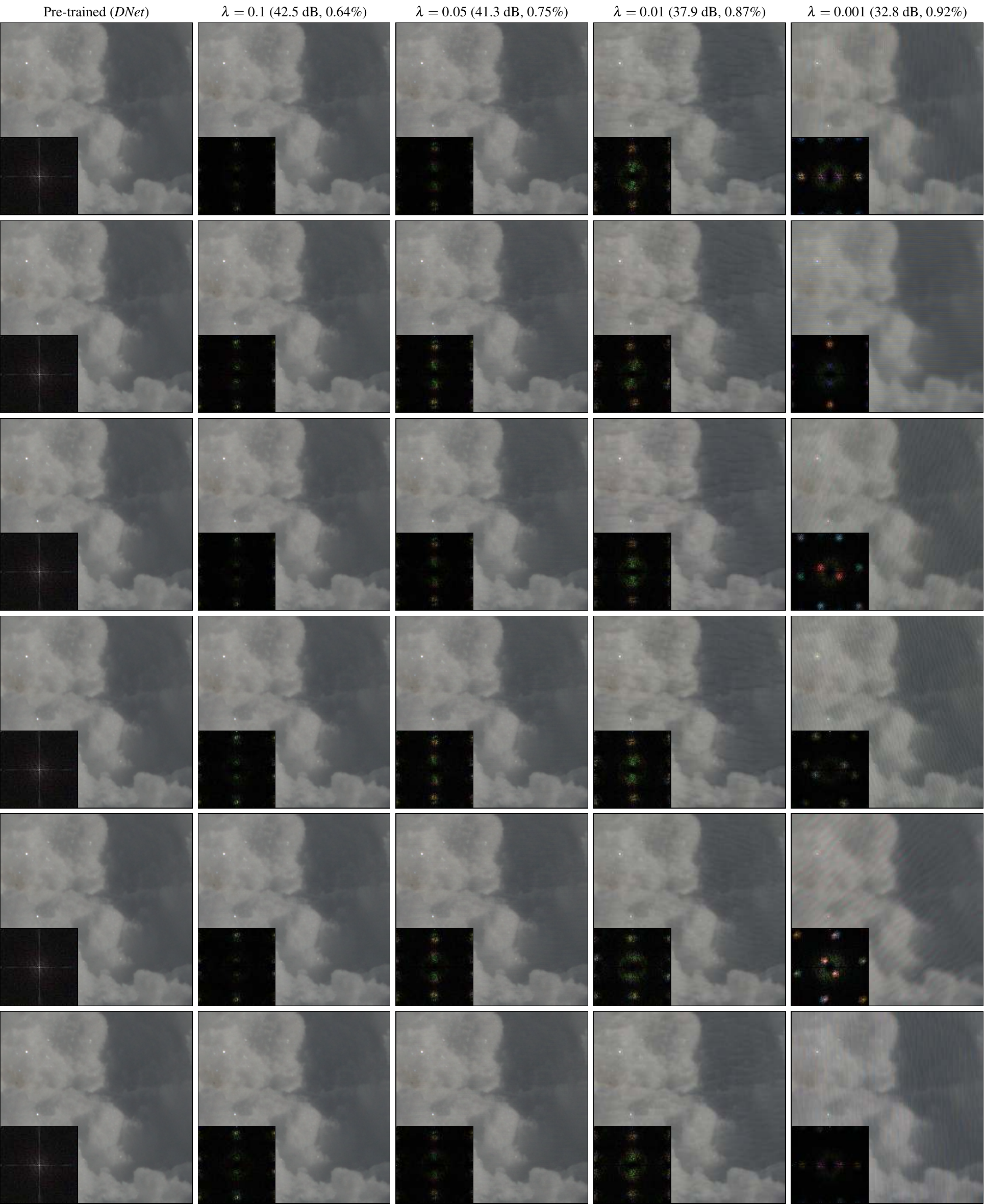}
    \caption{Variations in the learned distortions across training repetitions for various regularization strengths (\emph{DNet} model); reported validation metrics correspond to average values across all repetitions; for the pre-trained NIP the illustrated spectrum corresponds to the difference w.r.t. the optimization target; for the optimized NIPs, we show the difference between the spectra.}
    \label{fig:distortion-variations-dnet}
\end{figure*}

\begin{figure*}[p]
    \centering
    \includegraphics[width=0.90\textwidth]{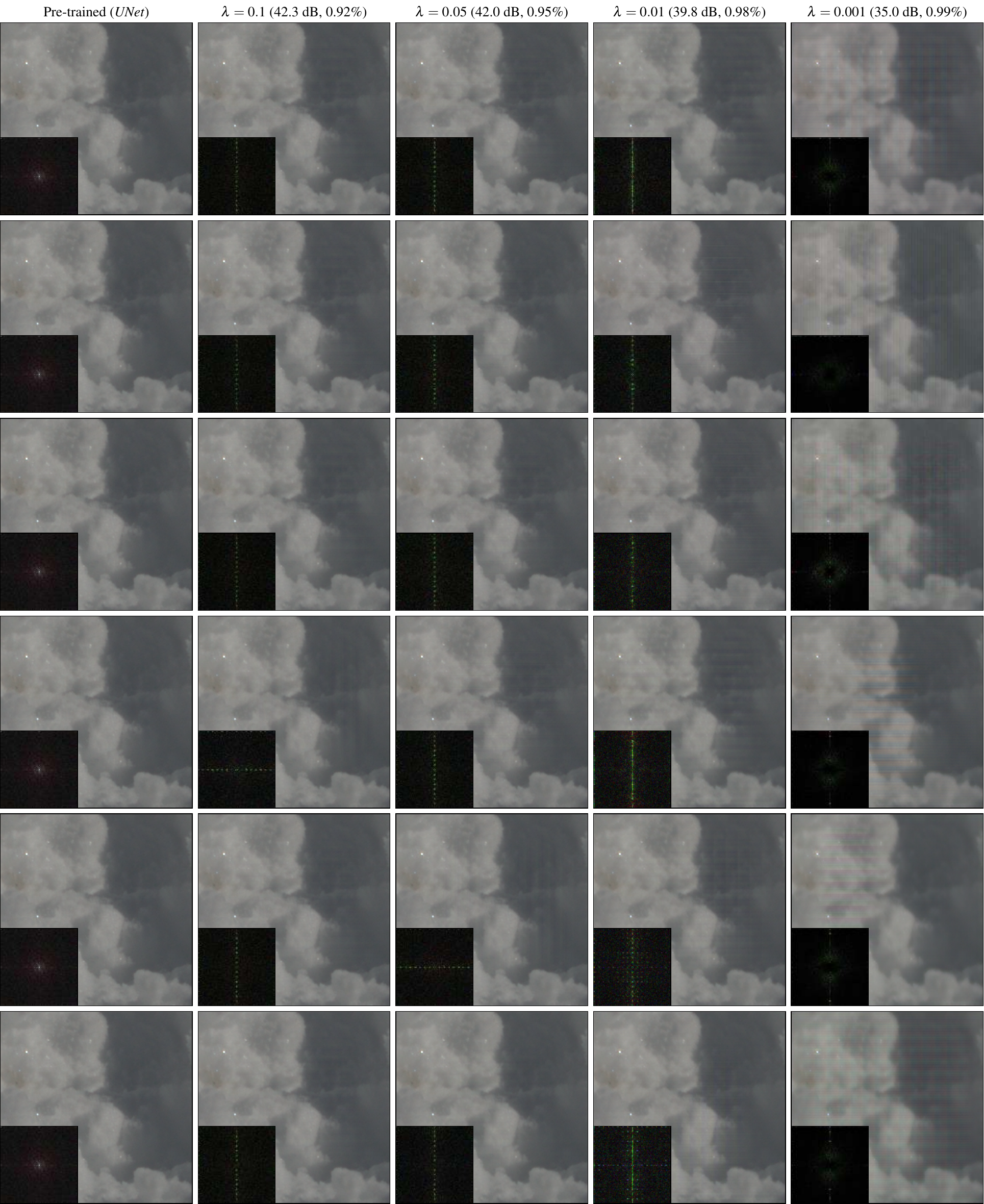}
    \caption{Variations in the learned distortions across training repetitions for various regularization strengths (\emph{UNet} model); reported validation metrics correspond to average values across all repetitions; for the pre-trained NIP the illustrated spectrum corresponds to the difference w.r.t. the optimization target; for the optimized NIPs, we show the difference between the spectra.}
    \label{fig:distortion-variations-unet}
\end{figure*}

\clearpage

\begin{table*}[p]
    \caption{The INet architecture: 321 trainable parameters}
    \label{tab:inet}
    \centering
    \begin{footnotesize}
\centering
\begin{tabular}{lllll}
    \toprule     
    \textbf{Operation} & \textbf{Activation} & \textbf{Initialization} & \textbf{Function} & \textbf{Output size} \tabularnewline
    \midrule     
    Input                      & - & - & RGGB feature maps & $N\times \frac{h}{2}\times \frac{w}{2} \times4$\tabularnewline
    $1 \times 1$ convolution   & - & hand-crafted binary sample selection$^{1}$ & Reorganizes data for up-sampling & $N\times \frac{h}{2}\times \frac{w}{2} \times 12 $\tabularnewline
    Depth to space              & - & - & Up-sampling & $N\times h\times w \times 3 $\tabularnewline
    $5 \times 5$ convolution   & - & zero-padded $3\times{}3$ bilinear kernel & Demosaicing & $N\times h\times w \times 3 $\tabularnewline
    $1 \times 1$ convolution   & - & sample color conversion matrix & Color-space conversion (sRGB) & $N\times h \times w \times 3 $\tabularnewline
    $1 \times 1$ convolution   & tanh & pre-trained model & Gamma correction$^{2}$ & $N\times h \times w \times 12 $\tabularnewline
    $1 \times 1$ convolution   & - & pre-trained model & Gamma correction$^{2}$ & $N\times h \times w \times 3 $\tabularnewline
    Clip to [0,1]              & - & - & output RGB image & $N\times h \times w \times 3 $\tabularnewline
    \bottomrule
    \multicolumn{5}{l}{$^{1}$ we disabled optimization of this filter to speed up convergence} \tabularnewline
    \multicolumn{5}{l}{$^{2}$ adapted from a 2-layer network trained separately to approximate gamma correction} \tabularnewline

    \end{tabular}    
\end{footnotesize}    

\end{table*}

\begin{table*}[p]
    \caption{The UNet architecture: 7,760,268 trainable parameters}
    \label{tab:unet}
    \centering
    \begin{footnotesize}
\centering
\begin{tabular}{lllll}
    \toprule     
    \textbf{Operation} & \textbf{Activation} & \textbf{Input} & \textbf{Output} & \textbf{Output size} \tabularnewline
    \midrule     
    Input                      & - & - & $x$ & $N\times h/2 \times w/2 \times4 $\tabularnewline
    $3 \times 3$ convolution   & leaky ReLU & $x$ & $c_{1,1}$ & $N\times h/2 \times w/2 \times 32 $ \tabularnewline
    $3 \times 3$ convolution   & leaky ReLU & $c_{1,1}$ & $c_{1,2}$ & $N\times h/2 \times w/2 \times 32 $ \tabularnewline
    $2 \times 2$ max pooling   & - & $c_{1,2}$ & $p_1$ & $N\times h/4 \times w/4 \times 32 $ \tabularnewline
    \midrule
    $3 \times 3$ convolution   & leaky ReLU & $p_1$ & $c_{2,1}$ & $N\times h/4 \times w/4 \times 64 $ \tabularnewline
    $3 \times 3$ convolution   & leaky ReLU & $c_{2,1}$ & $c_{2,2}$ & $N\times h/4 \times w/4 \times 64 $ \tabularnewline
    $2 \times 2$ max pooling   & - & $c_{2,2}$ & $p_2$ & $N\times h/8 \times w/8 \times 32 $ \tabularnewline
    \midrule
    $3 \times 3$ convolution   & leaky ReLU & $p_2$ & $c_{3,1}$ & $N\times h/8 \times w/8 \times 128 $ \tabularnewline
    $3 \times 3$ convolution   & leaky ReLU & $c_{3,1}$ & $c_{3,2}$ & $N\times h/8 \times w/8 \times 128 $ \tabularnewline
    $2 \times 2$ max pooling   & - & $c_{3,2}$ & $p_3$ & $N\times h/16 \times w/16 \times 128 $ \tabularnewline
    \midrule
    $3 \times 3$ convolution   & leaky ReLU & $p_3$ & $c_{4,1}$ & $N\times h/16 \times w/16 \times 256 $ \tabularnewline
    $3 \times 3$ convolution   & leaky ReLU & $c_{4,1}$ & $c_{4,2}$ & $N\times h/16 \times w/16 \times 256 $ \tabularnewline
    $2 \times 2$ max pooling   & - & $c_{4,2}$ & $p_4$ & $N\times h/32 \times w/32 \times 256 $ \tabularnewline
    \midrule
    $3 \times 3$ convolution   & leaky ReLU & $p_4$ & $c_{5,1}$ & $N\times h/32 \times w/32 \times 512 $ \tabularnewline
    $3 \times 3$ convolution   & leaky ReLU & $c_{5,1}$ & $c_{5,2}$ & $N\times h/32 \times w/32 \times 512 $ \tabularnewline
    $2 \times 2$ strided convolution & - & $c_{5,2}$ & $s_{5}$ & $N\times h/16 \times w/16 \times 256$ \tabularnewline
    \midrule
    $3 \times 3$ convolution   & leaky ReLU & $s_5~|~c_{4,2}$ & $c_{6,1}$ & $N\times h/16 \times w/16 \times 256 $ \tabularnewline
    $3 \times 3$ convolution   & leaky ReLU & $c_{6,1}$ & $c_{6,2}$ & $N\times h/16 \times w/16 \times 256  $ \tabularnewline
    $2 \times 2$ strided convolution & - & $c_{6,2}$ & $s_{6}$ & $N\times h/8 \times w/8 \times 128$ \tabularnewline
    \midrule
    $3 \times 3$ convolution   & leaky ReLU & $s_6~|~c_{3,2}$ & $c_{7,1}$ & $N\times h/8 \times w/8 \times 128 $\tabularnewline
    $3 \times 3$ convolution   & leaky ReLU & $c_{7,1}$ & $c_{7,2}$ & $N\times h/8 \times w/8 \times 128 $\tabularnewline
    $2 \times 2$ strided convolution & - & $c_{7,2}$ & $s_{7}$ & $N\times h/4 \times w/4 \times 64$ \tabularnewline
    \midrule
    $3 \times 3$ convolution   & leaky ReLU & $s_7~|~c_{2,2}$ & $c_{8,1}$ & $N\times h/4 \times w/4 \times 64 $\tabularnewline
    $3 \times 3$ convolution   & leaky ReLU & $c_{8,1}$ & $c_{8,2}$ & $N\times h/4 \times w/4 \times 64 $\tabularnewline
    $2 \times 2$ strided convolution & - & $c_{7,2}$ & $s_{8}$ & $N\times h/2 \times w/2 \times 32$ \tabularnewline
    \midrule
    $3 \times 3$ convolution   & leaky ReLU & $s_8~|~c_{1,2}$ & $c_{9,1}$ & $N\times h/2 \times w/2 \times 32 $\tabularnewline
    $3 \times 3$ convolution   & leaky ReLU & $c_{9,1}$ & $c_{9,2}$ & $N\times h/2 \times w/2 \times 32 $\tabularnewline
    $1 \times 1$ convolution   & - & $c_{9,2}$ & $c_{10}$ & $N\times h/2 \times w/2 \times 12 $\tabularnewline
    \midrule
    Depth to space   & - & $c_{10}$ & $y_{\text{rgb}}$ & $N\times h \times w \times 3 $\tabularnewline
    Clip to [0,1]    & - & $y_{\text{rgb}}$ & $y$ & $N\times h \times w \times 3 $\tabularnewline
    \bottomrule
    \multicolumn{4}{l}{All leaky ReLUs have $\alpha=0.2$} \tabularnewline
    \multicolumn{4}{l}{$|$ denotes concatenation along the feature dimension} \tabularnewline

    \end{tabular}    
\end{footnotesize}    

\end{table*}

\begin{table*}[p]
    \caption{The DNet architecture: 493,976 trainable parameters}
    \label{tab:dnet}
    \centering
    \begin{footnotesize}
\centering
\begin{tabular}{lllll}
    \toprule     
    \textbf{Operation} & \textbf{Activation} & \textbf{Input} & \textbf{Output} & \textbf{Output size} \tabularnewline
    \midrule     
    Input                      & - & - & $c_0$ & $N\times h/2 \times w/2 \times4 $\tabularnewline
    \midrule
    \multicolumn{4}{l}{Repeat for $i$ = 1, 2, \ldots, 14 \{} \tabularnewline
    ~~~~$3 \times 3$ convolution + BN  & ReLU & $c_{i-1}$ & $\hat{c}_{i}$  & $N\times h/2 - 2 \times w/2 - 2 \times 64 $ \tabularnewline
    ~~~~Padding (reflection)       & - & $\hat{c}_{i}$   & $c_{i}$  & $N\times h/2 \times w/2 \times 64 $ \tabularnewline
    \multicolumn{4}{l}{\}} \tabularnewline
    \midrule
    $3 \times 3$ convolution + BN   & ReLU & $c_{14}$  & $\hat{c}_{15}$  & $N\times h/2 - 2 \times w/2 - 2 \times 12 $ \tabularnewline
    Padding (reflection)       & - & $\hat{c}_{15}$  & $c_{15}$  & $N\times h/2 \times w/2 \times 12 $ \tabularnewline
    \midrule
    Depth to space             & - & $c_{15}$   & $f_\text{conv}$    & $N\times h \times w \times 3 $ \tabularnewline
    \midrule
    $1 \times 1$ convolution   & - & $c_{0}$   & $c_{16}$    & $N\times h/2 \times w/2 \times 12 $ \tabularnewline
    Depth to space             & - & $c_{16}$  & $f_\text{bayer}$    & $N\times h \times w \times 3 $ \tabularnewline
    \midrule
    $3 \times 3$ convolution   & ReLU & $f_\text{conv}~|~f_\text{bayer}$ & $\hat{c}_{17}$ & $N\times h - 2 \times w - 2 \times 64 $ \tabularnewline
    Padding (reflection)       & - & $\hat{c}_{17}$  & $c_{17}$  & $N\times h \times w \times 64 $ \tabularnewline
    $1 \times 1$ convolution   & - & $c_{17}$ & $y_{\text{rgb}}$ & $N\times h \times w \times 3 $ \tabularnewline
    Clip to [0,1]              & - & $y_{\text{rgb}}$ & $y$ & $N\times h \times w \times 3 $\tabularnewline
    \bottomrule
    \multicolumn{4}{l}{$|$ denotes concatenation along the feature dimension} \tabularnewline
    \end{tabular}    
\end{footnotesize}    

\end{table*}

\begin{table*}
    \caption{The dJPEG architecture for JPEG codec approximation}
    \label{tab:jpegnet}
    \centering
    \begin{footnotesize}
\centering
\begin{tabular}{llc}
    \toprule 
    \textbf{Operation} & \textbf{JPEG Function} & \textbf{Output size}\tabularnewline
    \midrule
     
    Input & - & $N\times h\times w\times3$\tabularnewline
     
    $1\times1$ convolution & RGB $\rightarrow$ YCbCr & $N\times h\times w\times3$\tabularnewline
     
    Space to depth \& reshapes & Isolate $8\times8$ px blocks & $3N\times8\times8\times B$\tabularnewline
     
    Transpose \& reshape & - & $3BN\times8\times8$\tabularnewline
     
    2 $\times$ matrix multiplication & Forward 2D DCT & $3BN\times8\times8$\tabularnewline
     
    Element-wise matrix division & Divide by quantization matrices & $3BN\times8\times8$\tabularnewline
     
    Rounding / approximate rounding & Quantization & $3BN\times8\times8$\tabularnewline
     
    Element-wise matrix multiplication & Multiply by quantization matrices & $3BN\times8\times8$\tabularnewline
     
    2 $\times$ matrix multiplication & Inverse 2D DCT & $3BN\times8\times8$\tabularnewline
     
    Transpose \& reshape & - & $3N\times8\times8\times B$\tabularnewline
     
    Depth to space \& reshapes & Re-assemble $8\times8$ px blocks & $N\times h\times w\times3$\tabularnewline
     
    $1\times1$ convolution & YCbCr $\rightarrow$ RGB & $N\times h\times w\times3$\tabularnewline
    \bottomrule
    \end{tabular}
\end{footnotesize}    

\end{table*}

\begin{table*}
    \caption{The FAN architecture: 1,341,990 trainable parameters}
    \label{tab:fan}
    \centering
    \begin{footnotesize}
\centering
\begin{tabular}{lllll}
    \toprule     
    \textbf{Operation} & \textbf{Activation} & \textbf{Initialization} & \textbf{Comment} & \textbf{Output size} \tabularnewline
    \midrule     
    Input                      & - & -    & RGB input & $N\times h\times w \times 3$\tabularnewline
    $5 \times 5$ convolution   & - & Standard residual filter$^1$ & Constrained convolution  & $N\times h \times w \times 3 $\tabularnewline
    \midrule
    $5 \times 5$ convolution   & leaky ReLU & MSRA & - & $N\times h \times w \times 32 $\tabularnewline
    $2 \times 2$ max pool      & - & -    & - & $N\times h/2 \times w/2 \times 32 $\tabularnewline
    $5 \times 5$ convolution   & leaky ReLU & MSRA & - & $N\times h/2 \times w/2 \times 64 $\tabularnewline
    $2 \times 2$ max pool      & - & -    & - & $N\times h/4 \times w/4 \times 64 $\tabularnewline
    $5 \times 5$ convolution   & leaky ReLU & MSRA & - & $N\times h/4 \times w/4 \times 128 $\tabularnewline
    $2 \times 2$ max pool      & - & -    & - & $N\times h/8 \times w/8 \times 128 $\tabularnewline
    $5 \times 5$ convolution   & leaky ReLU & MSRA & - & $N\times h/8 \times w/8 \times 256 $\tabularnewline
    $2 \times 2$ max pool      & - & -    & - & $N\times h/16 \times w/16 \times 256 $\tabularnewline
    \midrule
    $1 \times 1$ convolution   & leaky ReLU & MSRA & - & $N\times h/16 \times w/16 \times 256 $\tabularnewline
    \midrule
    global average pooling     & - & -    & - & $N \times 256 $\tabularnewline
    fully connected            & leaky ReLU & MSRA & - & $N\times 512 $\tabularnewline
    fully connected            & leaky ReLU & MSRA & - & $N\times 128 $\tabularnewline
    fully connected            & Softmax & MSRA & Class probabilities & $N\times 5 $\tabularnewline

    \bottomrule

    \end{tabular}    
\end{footnotesize}    

\end{table*}